\documentclass{article}

\newif\ifsub
\subtrue 

\usepackage{watml}
\usepackage[accepted]{icml2024}

\usepackage{dsfont}
\usepackage{microtype}
\usepackage{subfig}
\usepackage{nicefrac}       
\usepackage{xurl}            

\icmltitlerunning{Disguised Copyright Infringement of Latent Diffusion Models}

\renewcommand{\epsilon}{\varepsilon}

\begin{document}

\twocolumn[
\icmltitle{Disguised Copyright Infringement of Latent Diffusion Models}

\icmlsetsymbol{equal}{*}
\begin{icmlauthorlist}
\icmlauthor{Yiwei Lu}{equal,UW,Vec}
\icmlauthor{Matthew Y.R. Yang}{equal,UW}
\icmlauthor{Zuoqiu Liu}{equal,UW}
\icmlauthor{Gautam Kamath}{UW,Vec}
\icmlauthor{Yaoliang Yu}{UW,Vec}
\end{icmlauthorlist}

\icmlaffiliation{UW}{School of Computer Science, University of Waterloo, Waterloo, Canada}
\icmlaffiliation{Vec}{Vector Institute}

\icmlcorrespondingauthor{Yiwei Lu}{yiwei.lu@uwaterloo.ca}

\icmlkeywords{Machine Learning, ICML}

\vskip 0.3in
]
\printAffiliationsAndNotice{\icmlEqualContribution} 

\begin{abstract}
Copyright infringement may occur when a generative model produces samples substantially similar to some copyrighted data that it had access to during the training phase. The notion of access usually refers to including copyrighted samples \emph{directly} in the training dataset, which one may inspect to identify an infringement. We argue that such visual auditing largely overlooks a concealed copyright infringement, where one constructs a disguise that looks drastically different from the copyrighted sample yet still induces the effect of training Latent Diffusion Models on it. Such disguises only require \emph{indirect access} to the copyrighted material and cannot be visually distinguished, thus easily circumventing the current auditing tools. In this paper, we provide a better understanding of such disguised copyright infringement by uncovering the disguises generation algorithm, the revelation of the disguises, and importantly, how to detect them to augment the existing toolbox. Additionally, we introduce a broader notion of \emph{acknowledgment} for comprehending such \emph{indirect access}. Our code is available at \url{https://github.com/watml/disguised_copyright_infringement}.

\end{abstract}

\section{Introduction}

\begin{figure*}
    \centering
    \includegraphics[width=1.0\textwidth]{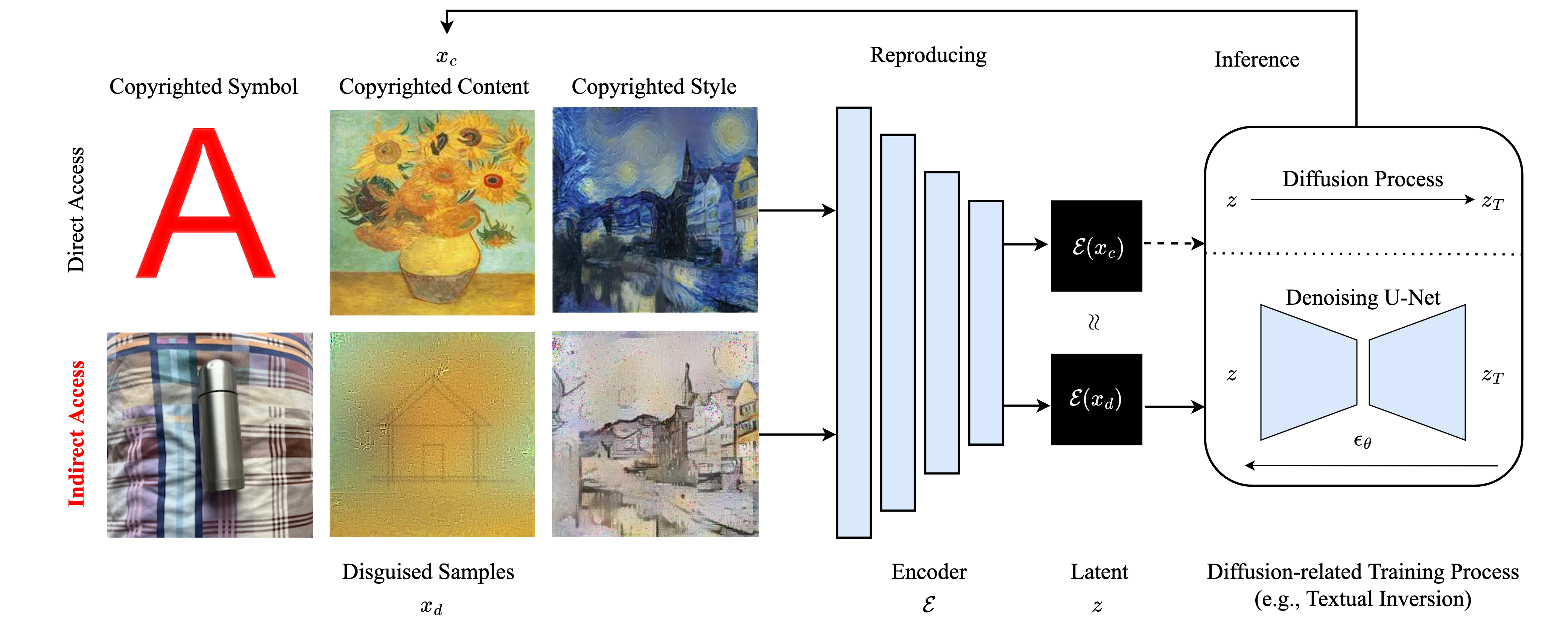}
    \caption{An overview of conventional (with \emph{direct access} to copyrighted material) and disguised (with \emph{indirect access}) copyright infringement for latent diffusion models. For \emph{direct access}, training an LDM-based model on copyrighted material $x_c$ and reproducing $x_c$ is subjected to copyright infringement. For \emph{indirect access}, one trains the same model on disguised samples $x_d$, which are drastically different from $x_c$, but is still able to reproduce $x_c$ during inference.    }
    \label{fig:intro}
    \vspace{-1em}
\end{figure*}

Generative models, especially the recent advanced Latent Diffusion Models (LDM) \cite{RombachBLEO22}, have shown tremendous ability to generate new images, even of creative or artistic form according to text prompts. Such models are trained on a large corpus of data, which may consist of copyrighted material. Additionally, prior works have established that such generative models are prone to regurgitating content from their training data \cite{IppolitoTNZJLCC23, ZhangILJTC21, CarliniIJLTZ22, VyasKB23,SimepalliSGGG23a,SimepalliSGGG23b}, which may also be copyrighted. 

In this paper, we will focus on copyright law within the jurisdiction of the United States.
To establish a copyright violation, two factors \emph{must} be present.
First, the accused must have had access to the copyrighted material. 
Second, the accused must produce content that bears ``substantial similarity'' to the copyrighted material (reproducing). 
Note that the definition of substantial similarity can be ambiguous. Within the context of images, its definition appears to be relatively broad \cite{1987steinberg}, and in particular encompasses near-exact copies.

Turning our attention to the former ``access'' criterion: the natural way to establish that a model had access to a particular piece of copyrighted material is to inspect its training data. 
For example, in the case of Andersen v.\ Stability AI Ltd.~\cite{2023andersen}, the case was allowed to proceed based on the fact that copyrighted images were found in LAION-5B~\cite{SchuhmannBVGWCCKMW22} (the training data used for Stable Diffusion) using haveibeentrained.com.

We challenge the perspective that such visual auditing is \emph{sufficient} to establish access to copyrighted material.
Our results show that it is possible to \emph{conceal} copyrighted images within the training dataset for LDMs. Specifically, LDMs are equipped with a fixed encoder for dimension reduction such that the diffusion learning process occurs in the latent space. This structure can be maliciously exploited to generate disguised copyrighted samples:  
given a copyrighted image, we show how to generate a disguise such that it is visually different from the copyrighted sample but shares similar latent information. The closeness of the two samples in the latent space can be quantitatively measured by a distance function, or qualitatively revealed by a concept extraction tool called textual inversion \cite{GalAAPBCC22}, both of which we will demonstrate in our empirical study.

Our study reveals the possibility of creating a new training dataset that does not appear to \emph{directly} or blatantly contain any copyrighted data. Nonetheless, if a model is trained on this derivative training dataset, it would behave similarly as if the copyrighted data were present. Such disguises may still exhibit copyright infringement, although only accessing proprietary data \emph{indirectly}. In \Cref{fig:intro}, we display a comparison between the previous copyright infringement phenomenon (\emph{direct access}) with the disguised copyright infringement (\emph{indirect access}). Clearly, there was still access to the copyrighted material in the latter training pipeline, which raises the  following question:

\begin{quotation}
\it What constitutes access? How to quantify it? 
\end{quotation}

We answer the first question by introducing a notion of \emph{acknowledgment}, which refers to a criterion that any sample that contains similar latent information as that of a copyrighted sample should be considered \emph{acknowledging} it, despite possible visual dissimilarity. To quantify \emph{acknowledgment} in practice, a deeper inspection than visually auditing the training set is required. Thus we further propose a two-step detection method: (1) a feature similarity search for screening suspects; (2) an encoder-decoder examination to confirm disguises, which augments the existing criterion. In summary, we make the following contributions:

\vspace{-0.5em}
\begin{itemize}
    \item We challenge the current ``access'' criterion and point out its insufficiency in more delicate cases of copyright infringement;
    \vspace{-0.5em}
    \item We propose an algorithm that demonstrably crafts disguised copyrighted data to conceal the content (or concepts) of copyrighted images in the training set;
    \vspace{-0.5em}
    \item  We show disguised data contain copyrighted information in the latent space, such that by finetuning them on textual inversion or DreamBooth, or training on LDM, the model reproduces copyrighted data during inference; 
    \vspace{-0.5em}
    \item We propose methods to detect such disguises, which further encourage the expansion and quantification of ``access'' in the context of copyright infringement.
\end{itemize}

\section{Background}

\subsection{Diffusion Models}

We focus on latent diffusion models \cite{RombachBLEO22} as they are ideal for text-to-image generation. We first recall the objective of regular diffusion models \cite{SohlWMG15}:
\begin{align}
    \mathcal{L}_{\texttt{DM}} = \mathbb{E}_{x,\epsilon,t} \Bigr[\|\epsilon-\epsilon_{\theta}(x_t,t)\|_2^2\Bigr],
\end{align}
where $\epsilon\sim\Nc(0,\mathbb{I})$, $t$ is the timestep uniformly sampled from $\{1, . . . , T \}$, and $x_t$ is a noisy version of the input sample $x$ at timestep $t$ . Briefly, diffusion models are probabilistic models designed to
learn a data distribution $p(x)$ by gradually denoising a normally distributed variable, which corresponds to learning
the reverse process of a fixed Markov chain of length $T$. 

Note that the intermediate $x_t$ are all in pixel space, which makes the training and inference of diffusion models expensive. To address this problem, \citet{RombachBLEO22} propose to perform the diffusion process in the latent space, namely latent diffusion models (LDM). Specifically, LDMs utilize a pre-trained (and fixed) autoencoder architecture which consists of an encoder $\mathcal{E}$ and a decoder $\mathcal{D}$, where $\mathcal{E}$ is only used during training and $\mathcal{D}$ is only used during inference. The objective of LDM can be expressed as:
\begin{align}
    \mathcal{L}_{\texttt{LDM}} = \mathbb{E}_{x,\epsilon,t} \Bigr[\|\epsilon-\epsilon_{\theta}(\mathcal{E}(x_t),t)\|_2^2\Bigr]. \label{eq:ldm}
\end{align}

When we perform text-to-image tasks, a text condition (or prompt) $y$ needs to be considered. Such a text condition is compressed into the latent space with a pre-trained text-embedding model $c_{\theta}(\cdot)$ (typically a BERT \cite{DevlinCLT18} text encoder) and fed into LDM training as a conditioning vector:
\begin{align}
    \mathcal{L}_{\texttt{LDM-c}} = \mathbb{E}_{x,y,\epsilon,t} \Bigr[\|\epsilon-\epsilon_{\theta}(\mathcal{E}(x_t),t, c_{\theta}(y))\|_2^2\Bigr]. \label{eq:ldm-c}
\end{align}

Intuitively, LDMs perform the diffusion training process on a  compressed version of the training data, while the compression procedure is deterministic. As a result, it is possible that two training samples share similar latent representations, but differ significantly visually. 
Based on this intuition, we propose a method to generate disguised data, demonstrating that the use of pre-trained autoencoders may \emph{conceal} the inclusion of copyrighted data in the training set.

\subsection{Data Poisoning attacks}

Data poisoning attacks refer to the threat of contaminating part of the training data such that a machine learning model trained on the mixture of clean and poisoned data is influenced toward certain behaviors. Existing data poisoning attacks mainly focus on poisoning supervised models (classifiers). Based on the objective, there exist (1) indiscriminate attacks \citep[\eg,][]{BiggioNL12,KL17,KohSL18,GonzalezBDPWLR17,LuKY22,LuKY23,LuYKY24}, which seek to decrease the overall test accuracy; (2) targeted attacks \citep[\eg,][]{ShafahiHNSSDG18,AghakhaniMWKV20,GuoL20,ZhuHLTSG19e,GeipingFHCTMG20} that only alter the prediction on specific test examples; (3) backdoor attacks \citep[\eg,][]{GuDG17,TranLM18,ChenLLLS17,SahaSP20} that trigger if a particular pattern appears in an example; (4) unlearnable examples \citep[\eg,][]{LiuC10,HuangMEBW21,YuZCYL21,FowlGCGCG21, FowlCGGBCG21,SadovalSGGGJ22,FuHLST21} that aim to reduce the utility of the data towards training a model. Moreover, the recent advanced Nightshade \cite{ShanDPZZ23} shows some success in poisoning LDMs on prompt-specific text-to-image tasks; \textcite{WanWSK23,JiangKZCB23} also explore data poisoning attacks against large language models.

Copyright infringement can be regarded as a special case of data poisoning, where one includes some ``poisoned data'' as a subset of a training set $\chi$, such that after training a LDM on $\chi$, the model reproduces a copyrighted sample $x_c$. For \emph{direct access}, ``poisoned data'' simply refers to direct copies of $x_c$. In this paper, we also examine \emph{indirect access}, where one can construct the ``poisoned data'' so they are visually different from $x_c$ using an adaptation of the data poisoning attack in \citet{ShafahiHNSSDG18}.

\subsection{Textual Inversion}

Art generation and related copyrighted material reproduction frequently happen in LDM-based generative tools. For example, textual inversion \cite{GalAAPBCC22} is a popular tool for modifying personalized images with language-guided LDMs. Textual inversion uses a small set of images (typically three to five), which depicts the target concept $S_*$ (which is a placeholder in a text prompt, e.g., ``A photo of $S_*$'') across multiple settings such as varied backgrounds or poses. The concept $S_*$ is tokenized , converted to an embedding $\nu_*$, and then encoded as part of the text embedding $c_{\theta}(y)$. The embedding $\nu_*$ is acquired by optimizing the LDM loss in Equation \ref{eq:ldm-c} over the training images:
\begin{align}
    \nu_* = \argmin_{\nu} \mathcal{L}_{\texttt{LDM-c}}
    \label{eq:textual_inversion}
\end{align}
Optimization is performed using the same training scheme as the original LDM while keeping $\epsilon_{\theta}$, $c_{\theta}$ and $\mathcal{E}$ fixed. After obtaining $\nu_*$, it can be combined with various text prompts (typically derived from the CLIP ImageNet templates \cite{RadfordKHRGASAMC21}) for generation. Overall, textual inversion is a great tool for capturing particular concepts and may also be used to reproduce copyrighted content. 
\section{Generating Disguises}

We describe how to \emph{conceal} copyrighted images within the training dataset. Specifically, we demonstrate that one can construct a disguised sample $x_d$ that is visually distinct from a copyrighted sample $x_c$, but contains essentially similar information in the latent space (measured via the distance between feature representations).
Specifically, we first recall the feature matching attack \cite{ShafahiHNSSDG18} designed for targeted data poisoning attacks, and then discuss how to adapt the attack to generate disguises.

\begin{figure}
    \centering  \includegraphics[width=0.5\textwidth]{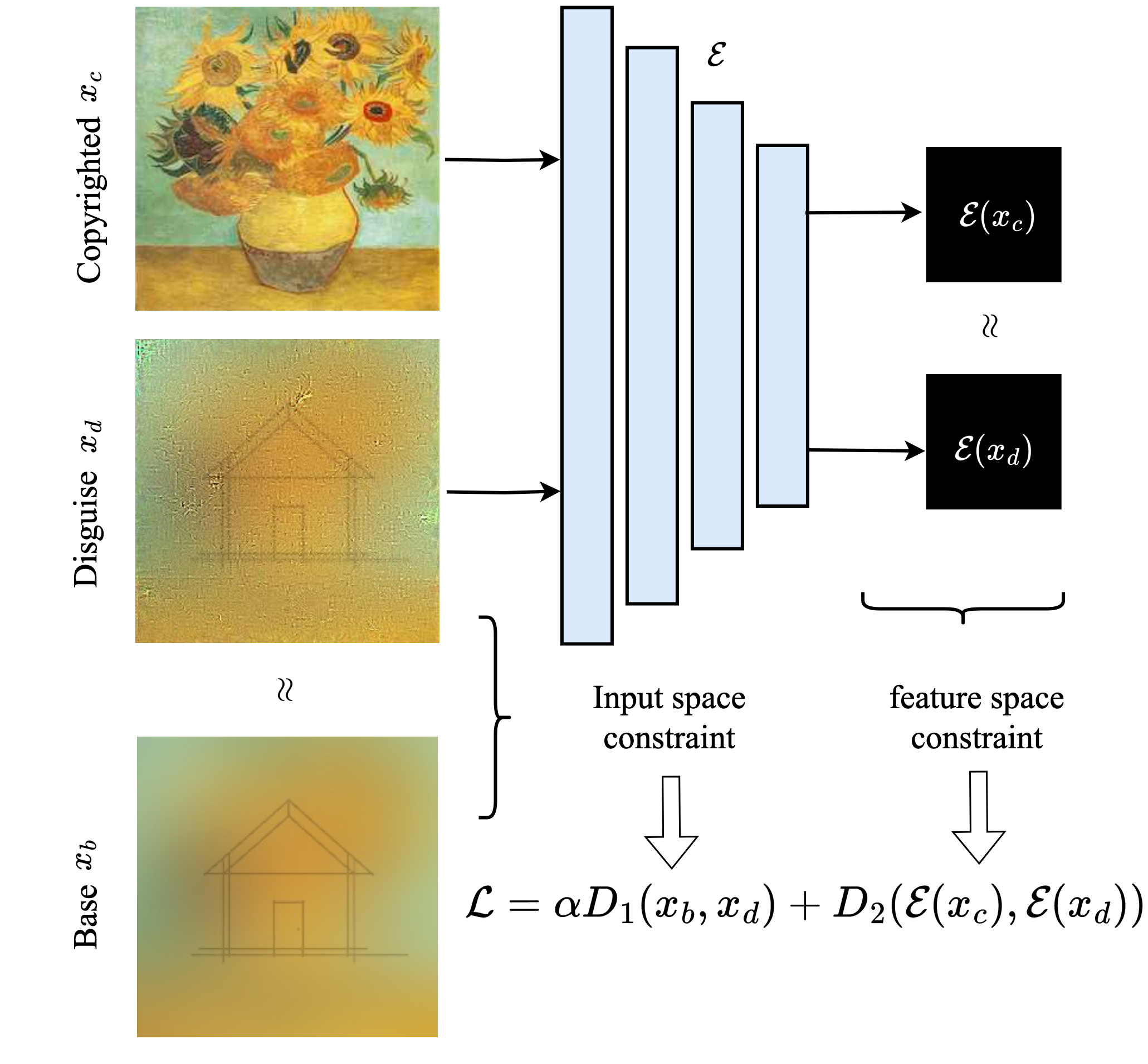}
    \caption{An illustration on the algorithm to generate disguises. We aim to optimize the loss $\mathcal{L}$ consisting of an input space constraint that measures the distance between the base image $x_b$ and the disguise $x_d$ in the input space, and a feature space constraint that measures the distance between the copyrighted $x_c$ and $x_d$ in the feature space. }
    \label{fig:method}
    \vspace{-1em}
\end{figure}

\subsection{Feature Matching Attack}

We recall that targeted attacks aim to change the prediction (e.g., causing misclassification) of a model (typically a classifier) on a targeted test sample $x_t$ (with label $y_t$) by training 
on a mixture of the existing clean data and poisoned data. Specifically,
\textcite{ShafahiHNSSDG18} propose a feature matching attack, where a poisoned sample $x_p$ is acquired by making imperceptible changes to a base sample $x_b$ with label $y_b$ (different than $y_t$) such that the feature representation of the poisoned sample $f(x_p)$ matches that of the target sample $f(x_t)$, where the model $f$ is a (usually fixed) pre-trained feature extractor. By training the model (e.g., with a softmax layer on top of $f$ ) on $x_p$, it is likely to misclassify the target sample $x_t$ as the wrong (base) label $y_b$.

Note that the generality of the feature matching attack is restricted in the context of classification tasks because it is only applicable when the feature extractor $f$ is fixed, which may only be applicable for fine-tuning or transfer learning. However, we will show how it can be used to craft disguised copyrighted images in the next paragraph.

\vspace{-1em}
\paragraph{Adapting to copyright infringement:} Training LDMs on a dataset $\chi$
amounts to performing a regular diffusion learning process on a latent dataset $\zeta$, where $\zeta=\mathcal{E}(\chi)$. 
Suppose there exists a set of $N$copyrighted data $\{x_c^i\}_{i=1}^N  \subseteq \chi$, which we aim to substitute with the same amount of disguised data $\{x_d^i\}_{i=1}^N$ such that the latent dataset $\zeta$ stays intact. Compared with data poisoning, the target sample $x_t$ is set to $x_c$, the feature extractor $f$ is set to $\mathcal{E}$, the base sample can be any uncopyrighted image different from $x_c$, the sample we aim to optimize is $x_d$, and the feature matching algorithm can be adapted. Conveniently, the encoder model $\mathcal{E}$ is pre-trained and its weights are fixed during the training of diffusion, thus making the above attack realistic.

\subsection{Disguised Copyrighted Image Generation}

Now we introduce our disguise generation algorithm for individual images. Formally, given a target copyrighted image $x_c$, a base image $x_b$ which is chosen to be visually different from $x_c$, distance measures $D_1(\cdot)$ and $D_2(\cdot)$ for input space and latent space respectively, and a fixed pre-trained encoder $\mathcal{E}$, we aim to construct a disguised image $x_d$ such that it satisfies:
\begin{align}
    D_1 = D_1(x_b,x_d)\leq\gamma_1,  D_2 = D_2(\mathcal{E}(x_c),\mathcal{E}(x_d))\leq\gamma_2,
\end{align}
where $\gamma_1$ is the input threshold that measures the visual distance of $x_d$ compared with $x_b$, $\gamma_2$ is the threshold that measures the feauture distance between $x_c$ and $x_d$ , which can be both determined empirically. 
We can then express the objective function as follows:
\begin{align}
    \argmin_{x_d} \alpha D_1 + D_2,
\end{align}
where $\alpha$ is a tunable hyperparameter for controlling the tradeoff. To illustrate more, the extreme case $\alpha=0$ refers to the scenario where there is no input space constraint on $x_d$ while $\alpha=\infty$ refers to no feature space constraint. Our algorithm is summarized in \Cref{alg:DG} and \Cref{fig:method}.

\begin{algorithm}[t]
\DontPrintSemicolon
    \KwIn{copyrighted image $x_c$, base image $x_b$, pre-trained encoder $\mathcal{E}$, input threshold $\gamma_1$, feature threshold $\gamma_2$, distance measure on input space $D_1(\cdot)$, distance measure on feature space $D_2(\cdot)$, hyperparameter on input space constraint $\alpha$, learning rate $\eta$.}
    Initialize disguise $x_d$ with base image $x_b$
        
    \Repeat{$D_1 \leq\gamma_1$ and  $D_2 \leq\gamma_2$}{

    $D_1 \gets D_1(x_b,x_d)$ \tcp*{image distance}

    $D_2 \gets D_2(\mathcal{E}(x_c),\mathcal{E}(x_d))$ \tcp*{feature distance}

    $\mathcal{L} \gets \alpha D_1 + D_2$ \tcp*{calculate loss}
    
    $x_d \gets x_d - \eta \frac{\partial\mathcal{L}}{\partial x_d} $ \tcp*{update disguise}

    $x_d \gets \mathrm{Proj}_{\Gamma}(x_d)$ \tcp*{project to admissible set}
    }

\textbf{return} disguise $x_d$
\caption{Disguise Generation}
\label{alg:DG}
\end{algorithm}

\vspace{-1em}
\paragraph{Connection to Nightshade:} \citet{ShafahiHNSSDG18} introduced the idea of attacking a model by manipulating training datapoints so that their feature representation resembles that of a different point. They applied this principle to targeted data poisoning attacks against classifiers, whereas \citet{ShanDPZZ23} (in their work ``Nightshade'') and we apply it to LDMs. To highlight some of the differences between our work and that of \citet{ShanDPZZ23}, there are crucial technical differences in how we instantiate this algorithm. Specifically,  Nightshade intends to \emph{stop} LDM from producing the image containing the correlated concept. In contrast, we aim to \emph{urge}
LDM to reproduce the target image (copyrighted image) through indirect access. Furthermore, the two works employ this algorithm towards very different ends. Nightshade aims to sabotage the connection between an image and its corresponding text prompt, thus requiring a pair of (image, text) as input. In contrast, our method only requires the form of images as input.

\section{Revealing Disguises}
\label{sec:exp}

\begin{figure}
    \centering
    \subfloat[$x_c$]{\includegraphics[width=0.096\textwidth]{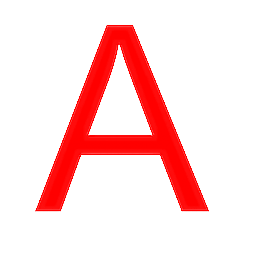}}
    \subfloat[Disguises $x_d$]{{\includegraphics[width=0.096\textwidth]{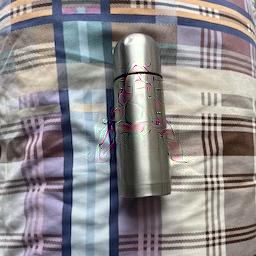}}
    {\includegraphics[width=0.096\textwidth]{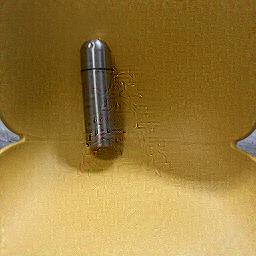}}
    {\includegraphics[width=0.096\textwidth]{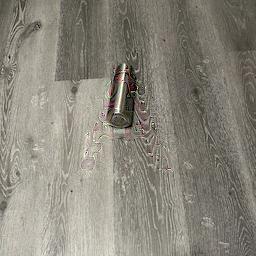}}
    {\includegraphics[width=0.096\textwidth]{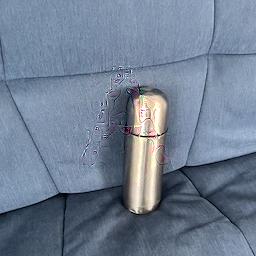}}}\\
    \vspace{-1em}
    \subfloat[Inference on textual inversion by learning on $x_d$ only]{\includegraphics[width=0.5\textwidth]{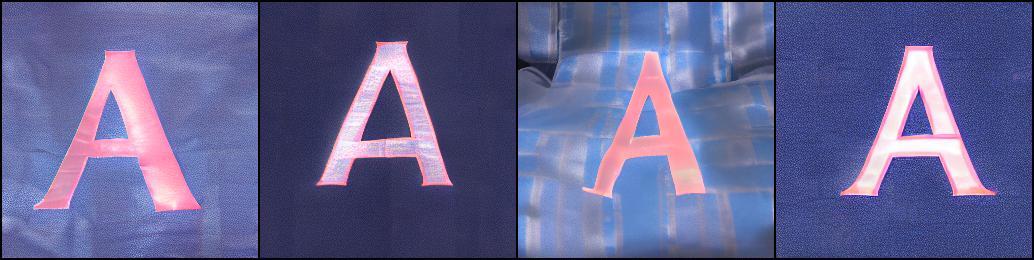}}

    \caption{The disguised symbol on textual inversion. The first row from left to right: Column (1) the designated copyrighted symbol; Columns (2)-(5) four disguises $x_d$ generated with different $x_b$. The second row: images generated by textual inversion after training on the above $x_d$.  }
    \label{fig:watermark}
    \vspace{-1em}
\end{figure}

\begin{figure}[ht]
    \centering
    {\includegraphics[width=0.15\textwidth]{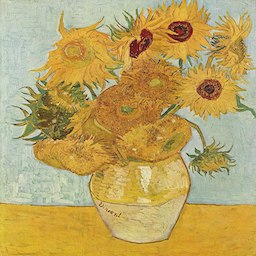}} 
    {\includegraphics[width=0.15\textwidth]{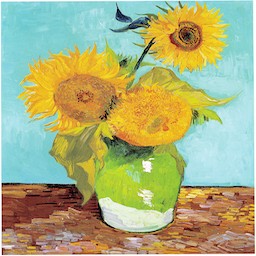}}
    {\includegraphics[width=0.15\textwidth]{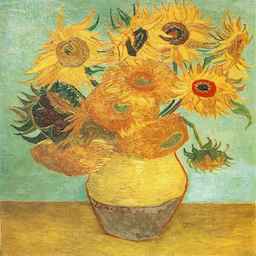}}\\
    \vskip 0.1cm
    {\includegraphics[width=0.15\textwidth]{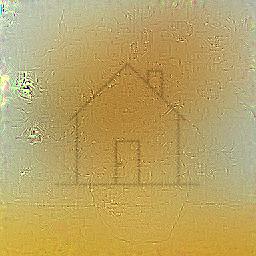}}
    {\includegraphics[width=0.15\textwidth]{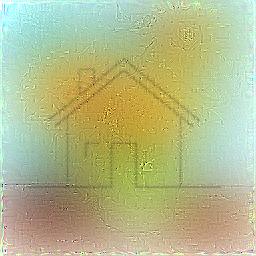}}
    {\includegraphics[width=0.15\textwidth]{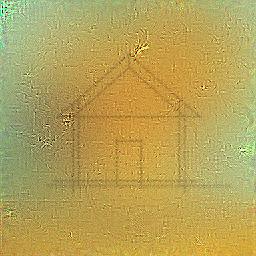}}\\
    \vskip 0.1cm
    
    {\includegraphics[width=0.46\textwidth]{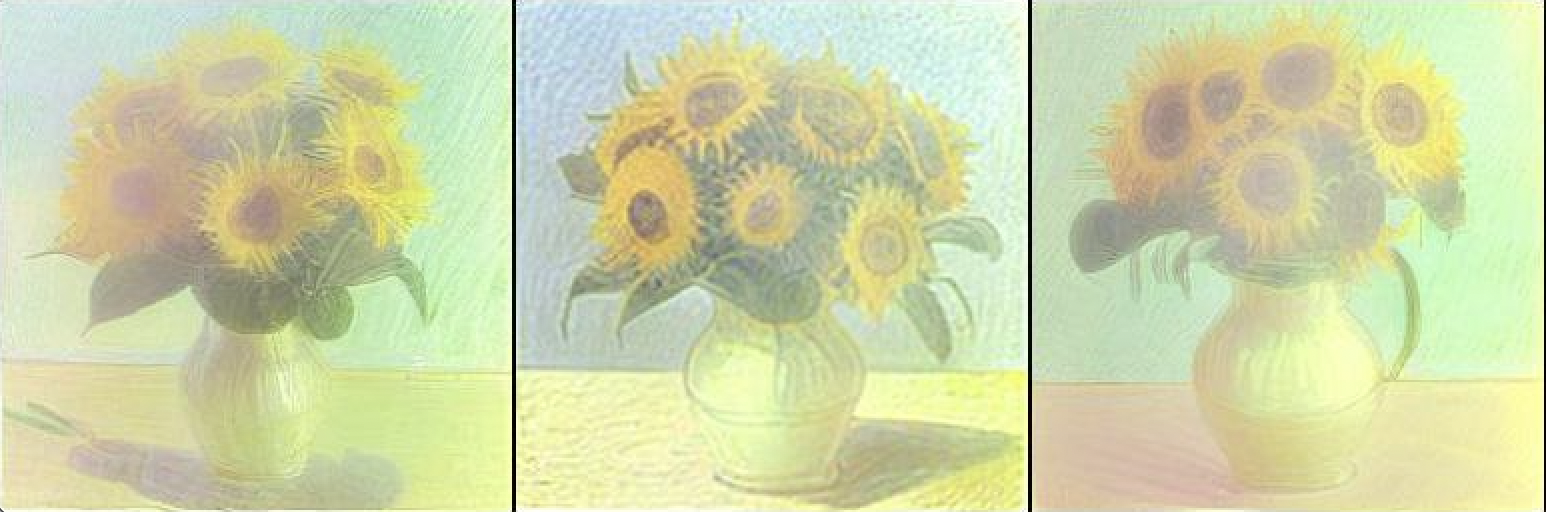}}

    \caption{We show the disguised copyrighted content on textual inversion. The first row: the designated copyrighted image $x_c$ (\emph{The Sunflowers} by Vincent Van Gogh); the second row: three disguises $x_d$ generated with different $x_b$; the third row: images generated by textual inversion after training on the above $x_d$.}
    \label{fig:objective}
    \vspace{-1em}
\end{figure}

\begin{figure}
    \centering
    {\includegraphics[width=0.15\textwidth]{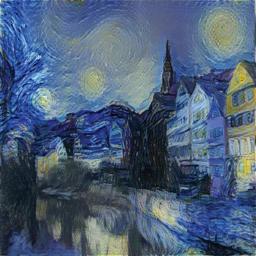}}
    {\includegraphics[width=0.15\textwidth]{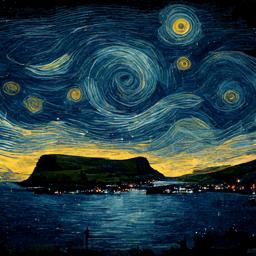}}
    {\includegraphics[width=0.15\textwidth]{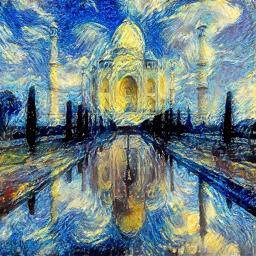}}\\
    \vskip 0.1cm
    {\includegraphics[width=0.15\textwidth]{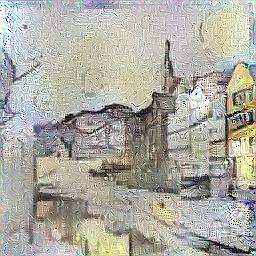}}
    {\includegraphics[width=0.15\textwidth]{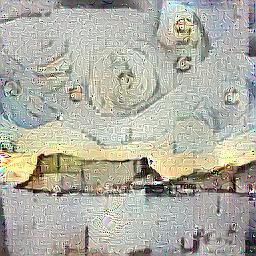}}
    {\includegraphics[width=0.15\textwidth]{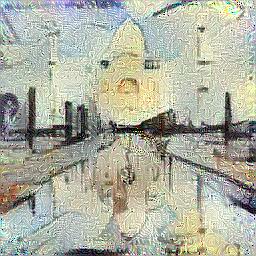}}\\
    \vskip 0.1cm
    {\includegraphics[width=0.46\textwidth]{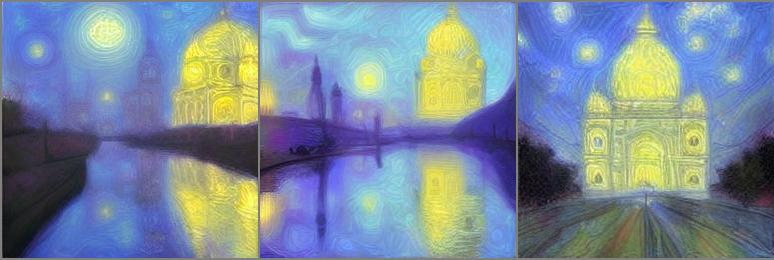}}

    \caption{We show the disguised copyrighted style on textual inversion. The first row: the designated copyrighted style $x_c$ (in the style of \emph{The Starry Night} by Vincent Van Gogh); the second row: disguises $x_d$ generated with different $x_b$; the third row: images generated by textual inversion after training on the above $x_d$.}
    \label{fig:style_main}
    \vspace{-1em}
\end{figure}

Recall that the set of disguised data $\{x_d^i\}_{i=1}^N$ is constructed to substitute their copyrighted counterpart $\{x_c^i\}_{i=1}^N$ such that the latent dataset $\zeta$ is not changed much. To evaluate whether this goal is accomplished, we extract the latent information contained in the acquired samples $x_d$ using textual inversion to qualitatively \emph{reveal} the disguises. We also report the feature threshold $\gamma_2$ for quantitative analysis.

Moreover, examination of textual inversion also demonstrates how disguised copyright infringement may occur in the wild, i.e., reproducing copyrighted materials using products that utilize pre-trained LDMs to generate arts, which bring major concerns for human artists. Such LDM-based methods usually require only a few images as the training set $\chi$, and can be easily substituted by disguises entirely. In \Cref{sec:dream_finetune}, we will further extend our evaluation to DreamBooth \citep{ruiz2023dreambooth} and LDM training.

\subsection{Experimental settings}
Note that the ``copyrighted'' images $x_c$ used in our experiments may not be actually copyrighted, but used as a substitute to demonstrate disguised copyright infringement.

\paragraph{LDM:} 

We adopt the official PyTorch implementation\footnote{\url{https://github.com/CompVis/latent-diffusion}} of conditional LDM \cite{RombachBLEO22} and acquire the pre-trained weights\footnote{\url{https://ommer-lab.com/files/latent-diffusion/nitro/txt2img-f8-large/model.ckpt}}(including that of the encoder $\mathcal{E}$, the denoising U-Net $\epsilon_{\theta}$ and the text embedding $c_{\theta}(\cdot)$) of a 1.45B parameter \emph{KL}-regularized LDM-$8$ ($8$ denotes downsampling factor) model conditioned on language prompts on LAION-400M \cite{SchuhmannKKKVBJCM21}. 
We apply the pre-trained encoder $\mathcal{E}$ in \Cref{alg:DG} to generate disguises, and the entire pre-trained LDM for textual inversion. Note that the LDM is not re-trained in our experiments.

\paragraph{Generating Disguises:} 

Throughout our experiments, we apply the pre-trained \emph{KL}-regularized encoder $\mathcal{E}$, and set the input distance measure $D_1(\cdot)$ as a sum of the multi-scale structural similarity index (MS-SSIM) loss \parencite{WangSB03} and $L_1$ loss following the analysis of  \cite{KTKO21}, and the feature distance measure to be the $L_2$ loss: $D_2(\mathcal{E}(x_c),\mathcal{E}(x_d))=\|\mathcal{E}(x_c) - \mathcal{E}(x_d)\|_2$. The choice of the copyrighted image $x_c$, base image $x_b$, the input threshold $\gamma_1$, the feature threshold $\gamma_2$ and the hyperparameter on the input constraint 
$\alpha$ are task-dependent (for different copyrighted material), and we specify them in their corresponding paragraphs below. We set the admissible set to be in the range of $[0,1]$ as the legitimate (normalized) image pixel value and run the algorithm for 100000 epochs (early stop if the stopping criteria are reached) for all experiments.

\paragraph{Textual inversion:} 

After acquiring the disguises $x_d$, we feed them (we generate $N=\{3,4\}$ disguises for extracting the same concept, following the normal recipe) into textual inversion \cite{GalAAPBCC22} and optimize the embedding $\nu_*$ with Equation \ref{eq:textual_inversion}. We adopt the official PyTorch implementation\footnote{\url{https://github.com/rinongal/textual_inversion}}, which utilizes the pre-trained \emph{KL}-regularized LDM-8 with the same encoder $\mathcal{E}$. Note that training textual inversion requires an initial word as input, which is an initialization for the concept $S_*$. Throughout our experiments, we choose the initial word to match the visual appearance of the disguises (thus the concept of the base image) and we observe that changing the initial word does not lead to a significant difference for the algorithm. For generating new images, we select from the pool of prompts included in the existing implementation of textual inversion\footnote{\url{https://github.com/rinongal/textual_inversion/blob/main/ldm/data/personalized.py}} which we specify for each task below.

\begin{figure*}[t]
    \centering
    
    {\includegraphics[width=0.16\textwidth]{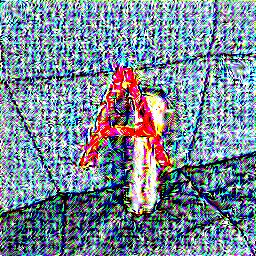}}
    {\includegraphics[width=0.16\textwidth]{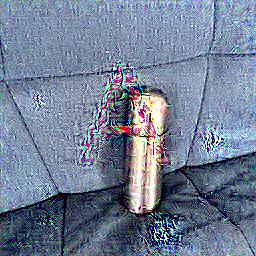}}
    {\includegraphics[width=0.16\textwidth]{images/watermark/bottle_clipped/poisoned4.jpg}}
    {\includegraphics[width=0.16\textwidth]{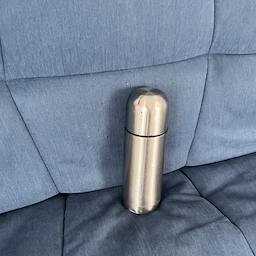}}
    {\includegraphics[width=0.16\textwidth]{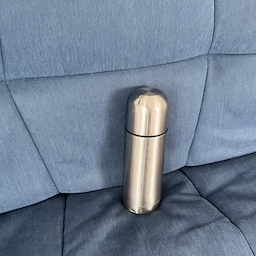}}
    {\includegraphics[width=0.16\textwidth]{images/watermark/bottle_clipped/target.png}}
    \vskip 0.1cm
    {\includegraphics[width=0.16\textwidth]{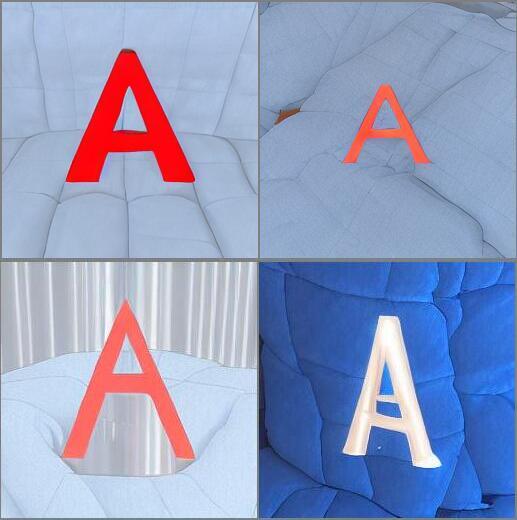}}
    {\includegraphics[width=0.16\textwidth]{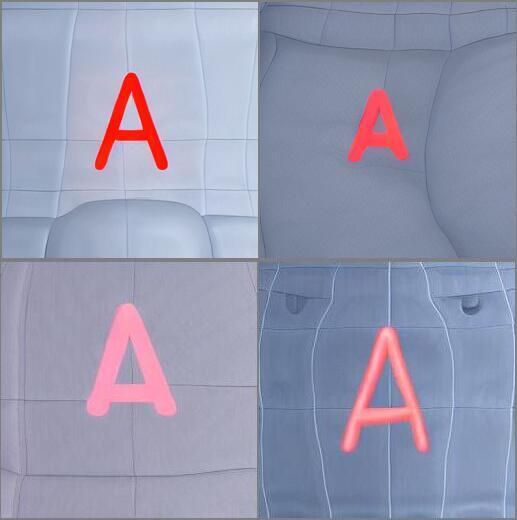}}
    {\includegraphics[width=0.16\textwidth]{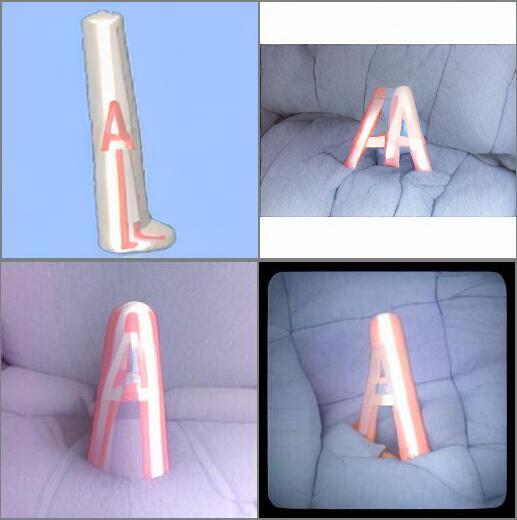}}
    {\includegraphics[width=0.16\textwidth]{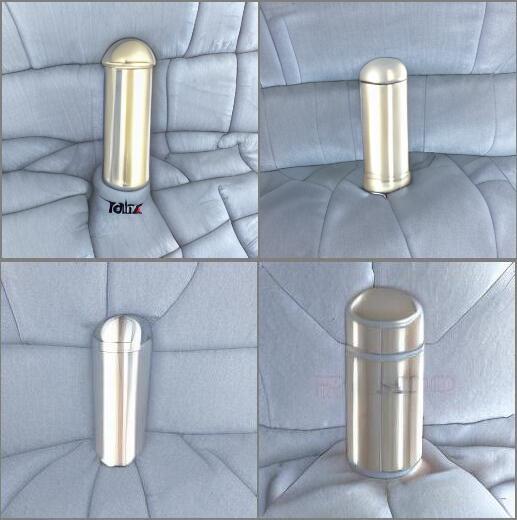}}
    {\includegraphics[width=0.16\textwidth]{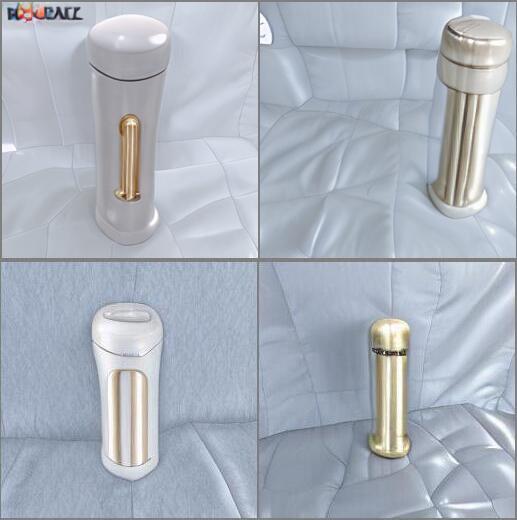}}
    {\includegraphics[width=0.16\textwidth]{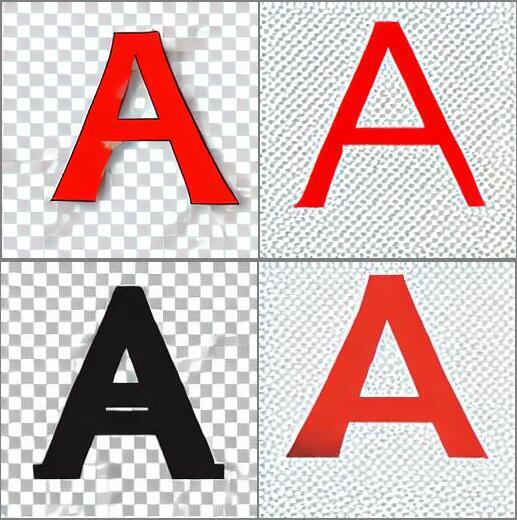}}
\vskip 0.1cm
    {\includegraphics[width=0.16\textwidth]{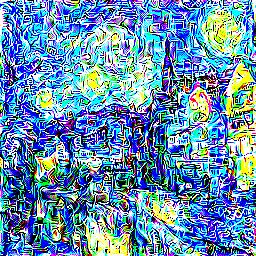}}
    {\includegraphics[width=0.16\textwidth]
    {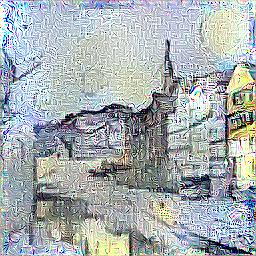}}
    {\includegraphics[width=0.16\textwidth]{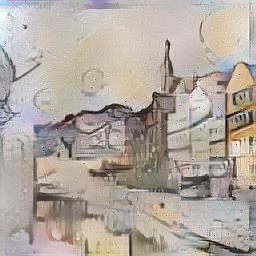}}
    {\includegraphics[width=0.16\textwidth]{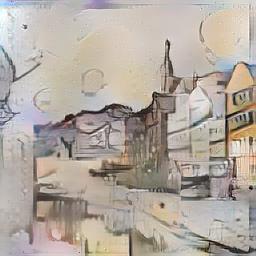}}
    {\includegraphics[width=0.16\textwidth]{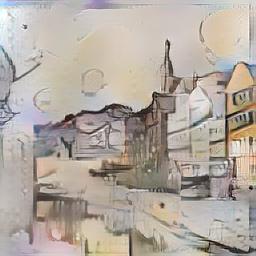}}
    {\includegraphics[width=0.16\textwidth]{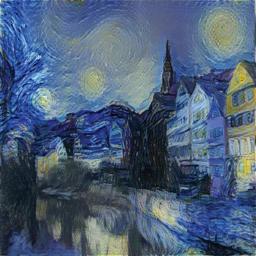}}
    \vskip 0.1cm
    \vspace{-1em}
    \subfloat[no constraint ($\alpha=0$)]{\includegraphics[width=0.16\textwidth]{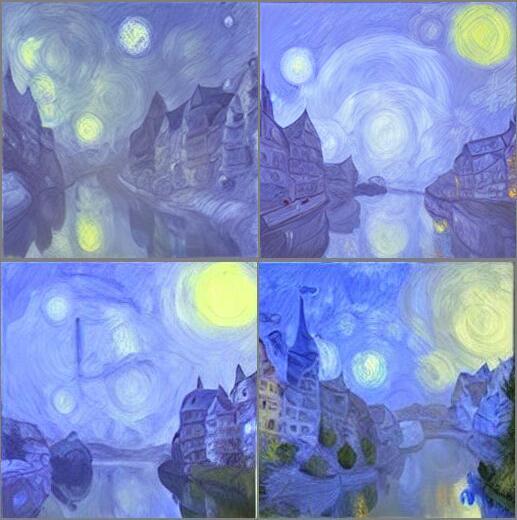}}
    \hskip 0.1cm
    \subfloat[$\alpha=1000$]{\includegraphics[width=0.16\textwidth]{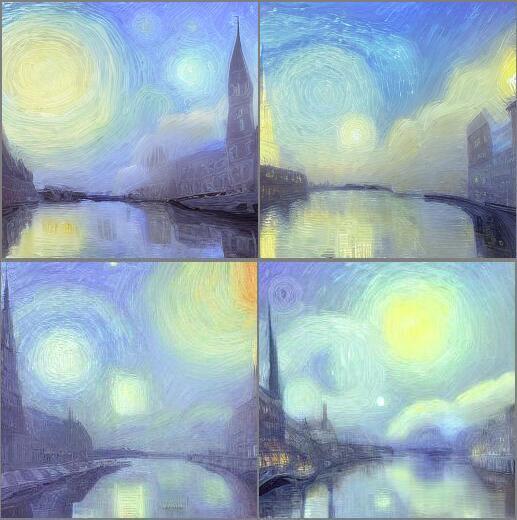}}
    \hskip 0.1cm
    \subfloat[$\alpha=8000$]{\includegraphics[width=0.16\textwidth]{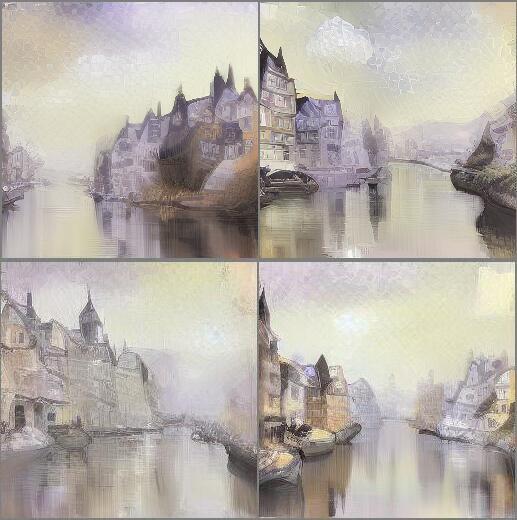}}
    \hskip 0.1cm
    \subfloat[$\alpha=64000$]{\includegraphics[width=0.16\textwidth]{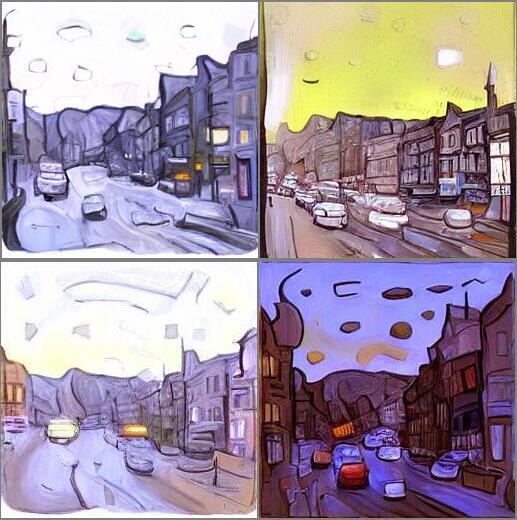}}
    \hskip 0.1cm
    \subfloat[base ($\alpha=\infty$)]{\includegraphics[width=0.16\textwidth]{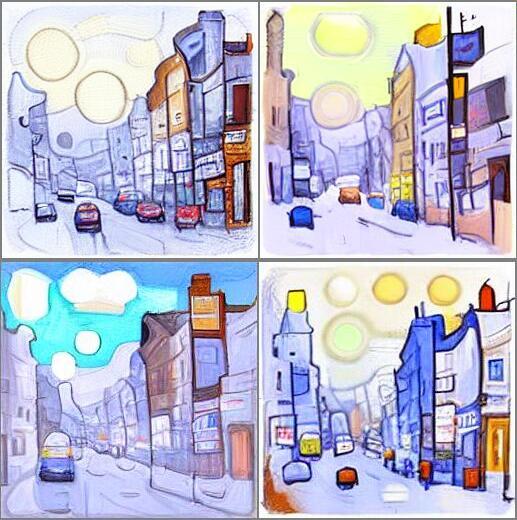}}
    \hskip 0.1cm
    \subfloat[target]{\includegraphics[width=0.16\textwidth]{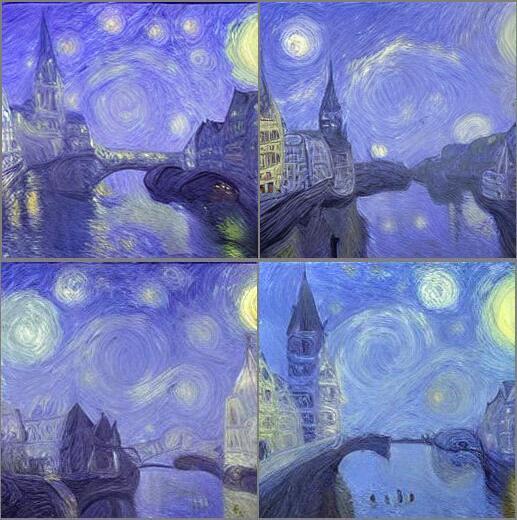}}

    \caption{The effect of tuning the hyperparameter $\alpha$ on disguises (Rows 1-2: symbol; Rows 3-4: style). Rows 1 and 3 show disguises generated with different $\alpha$; Rows 2 and 4 show the images generated by textual inversion by training on the disguises above. Note that the criterion on the input and feature threshold may not be satisfied in this series of experiments and we only train the disguise generation algorithm until convergence. }
\label{fig:noise_watermarking}
\vspace{-1em}
\end{figure*}

\begin{figure*}[t]
    \centering
    \subfloat[]{\includegraphics[width=0.11\textwidth]{images/watermark/bottle_clipped/poisoned1.jpg}}
    \subfloat[]{\includegraphics[width=0.11\textwidth]{images/watermark/bottle_clipped/poisoned2.jpg}}
    \subfloat[]{\includegraphics[width=0.11\textwidth]{images/watermark/bottle_clipped/poisoned3.jpg}}
    \subfloat[]{\includegraphics[width=0.11\textwidth]{images/entire_image/sunflower_clipped/poisoned5.jpg}}
    \subfloat[]{\includegraphics[width=0.11\textwidth]{images/entire_image/sunflower_clipped/poisoned6.jpg}}
    \subfloat[]{\includegraphics[width=0.11\textwidth]{images/entire_image/sunflower_clipped/poisoned2.jpg}}
    \subfloat[]{\includegraphics[width=0.11\textwidth]{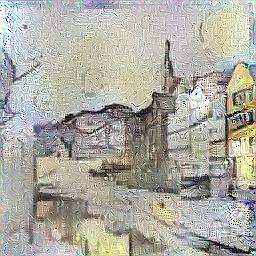}}
    \subfloat[]{\includegraphics[width=0.11\textwidth]{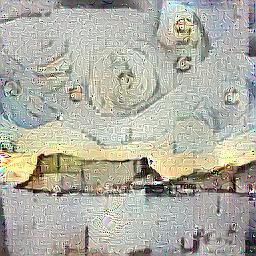}}
    \subfloat[]{\includegraphics[width=0.11\textwidth]{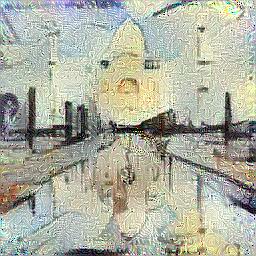}}
    \vspace{-2em}
    \subfloat[]{\includegraphics[width=0.11\textwidth]{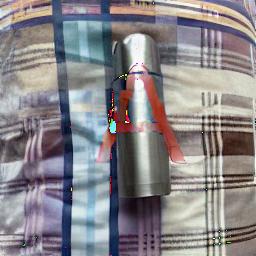}}
    \subfloat[]{\includegraphics[width=0.11\textwidth]{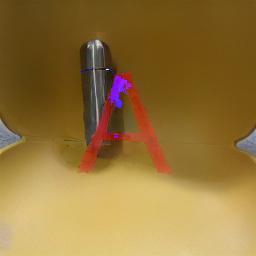}}
    \subfloat[]{\includegraphics[width=0.11\textwidth]{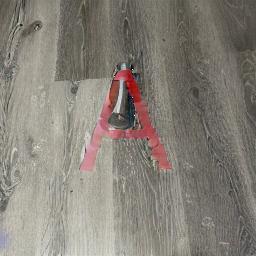}}
    \subfloat[]{\includegraphics[width=0.11\textwidth]{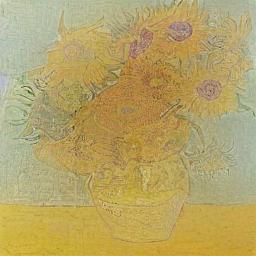}}
    \subfloat[]{\includegraphics[width=0.11\textwidth]{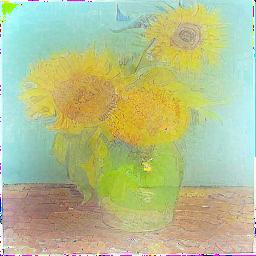}}
    \subfloat[]{\includegraphics[width=0.11\textwidth]{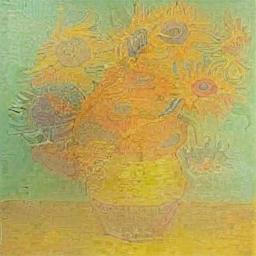}}
    \subfloat[]{\includegraphics[width=0.11\textwidth]{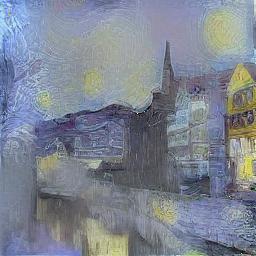}}
    \subfloat[]{\includegraphics[width=0.11\textwidth]{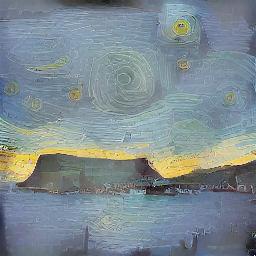}}
    \subfloat[]{\includegraphics[width=0.11\textwidth]{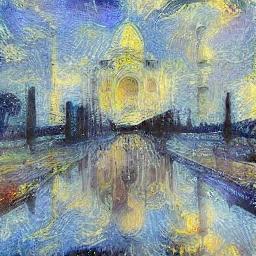}}
    
    \caption{Results for the encoder-decoder examination. The first row shows the disguises $x_d$ and the second row includes the output of the autoencoder: $\mathcal{D}(\mathcal{E}(x_d))$. Columns 1-3, 4-6, and 7-9 show results for symbol, content, and style, respectively.
    The autoencoder reveals the copyrighted information contained in the disguises.}
    \label{fig:detection}
\end{figure*}

\subsection{Concepts learned on disguises}

We demonstrate the concepts learned by textual inversion on the disguises with respect to different copyrighted patterns. 
(1) a copyrighted symbol; (2)  copyrighted content; and (3) style. We note that although the style is generally \emph{not} copyrightable, there are still ongoing debates about the \emph{ethics} of using copyrighted artwork to train a model that can imitate an artist's style\footnote{\url{https://news.bloomberglaw.com/ip-law/ai-imitating-artist-style-drives-call-to-rethink-copyright-law}}. For this reason, we also examine the case of hiding an artist's style in disguised images.

\paragraph{Disguised symbol:} We first show the easiest scenario to generate disguises, namely copyrighted symbol. In \Cref{fig:watermark}, we pick the symbol ``A'' as $x_c$ (note that here we abstract $x_c$ as the symbol ``A'', in our experiment, it is combined with each base image to create the corresponding copyrighted image $x_c$, see \Cref{fig:watermark_target} in \Cref{app:add_exp}) and choose four images of a water bottle as base images $x_b$. The base images do not need to be carefully constructed and are photos taken by ourselves. Each disguise $x_d$ is created by taking a pair of $x_c$ and $x_b$ as input to \Cref{alg:DG}. We set $\gamma_1 = 0.05,\gamma_2 = 0.35$ (normalized to the range of $[0,1]$) as the threshold of input and feature distance, respectively, and $\alpha=8000$. By feeding $x_d$ into textual inversion with the text prompt ``a photo of a *'', we reproduce the target symbol ``A'' without being exposed to the semantic information of the copyrighted content.

\paragraph{Disguised content}

We then show a more challenging task to generate disguised content. In \Cref{fig:objective}, we pick three drawings of \emph{The Sunflowers}\footnote{From left to right: F456 (\emph{1888}), F453 (\emph{1888}), F455 (\emph{1889}).} by Vincent Van Gogh as $x_c$ (first row). Due to the difficulty of the task, we cannot choose any image as $x_b$. Thus we first blur $x_c$ such that they lose their semantic information and retain the color pattern, then we add simple sketches of houses\footnote{These simple sketches of houses were generated by ChatGPT with the text prompt ``create a very very simple line art of a very simple house in the middle with a white background''} as base images $x_b$ (second row). In \Cref{fig:fail} in \Cref{app:add_exp}, we show the background is essential for our purpose, where the disguises are ineffective with a white background. We set $\gamma_1 = 0.08,\gamma_2 = 0.26, \alpha=4000$. By feeding $x_d$ into textual inversion with the text prompt ``a photo of a *'', we recover the content of the sunflowers (third row) without containing the semantic information of $x_d$.

\paragraph{Disguised style} Finally, we show the results for style scraping. In \Cref{fig:style_main}, we pick three drawings\footnote{From left to right: The Neckarfront in Tubingen, Germany (photo by Andreas Praefcke); The Faroe Islands (courtesy Listasavn Føroya); Taj Mahal (adamkaz/Getty Images) in the style of \emph{The Starry Night}, generated with Neural Style Transfer \cite{GatysEB15}.} in the style of \emph{The Starry Night} by Vincent Van Gogh as $x_c$ (first row). 
The base images $x_b$ (second row) are the target images with another style (watercolor), generated with AdaIN-based \cite{HuangB17} style transfer\footnote{\url{https://github.com/tyui592/AdaIN_Pytorch}}. We set $\gamma_1=0.03, \gamma_2=0.33$, $\alpha=2000$. By feeding $x_d$ into textual inversion with the text prompt  ``a painting in the style of *'', we reproduce the style of $x_c$ (\emph{The Starry Night}) while textual inversion only learns from images that visually resemble watercolor style.

In summary, for disguised symbols, generating the disguises $x_d$ with any base image $x_b$ is easy. However, for generating disguised content and style, one needs to choose the base image carefully for successful optimization.

\subsection{Visual appearance of disguises}

Next, we control the visual appearance of $x_d$ by tuning $\alpha$, which is the weighting parameter of the input space constraint.  
In \Cref{fig:noise_watermarking}, we show the tradeoff of tuning the hyperparameter $\alpha$:  (1) a smaller $\alpha$ indicates weaker input space constraint and it could lead to an ineffective disguise, where $x_d$ still visually contains the copyrighted material; (2) in contrast, a bigger $\alpha$ shifts the optimization focus away from feature matching to generate latent embeddings distinct from the copyrighted content's. In the latter case, the tradeoff is that despite having a strong disguise that visually hides the copyrighted content, the textual inversion process may not successfully learn the copyrighted material, thus rendering our attack pointless. Note that there is no input constraint for the extreme case $\alpha=0$, and the visual appearance of disguises largely depends on the initialization (we initialize with $x_b$ above) and we show the results for different initialization in \Cref{fig:init} in \Cref{app:add_exp}.

\subsection{Detection}

\label{sec:detection}

Next, we introduce a two-step detection method specifically for disguised samples that go beyond browsing through the training set, e.g., haveibeentrained.com.

(1) \textbf{Feature similarity search:}
Quantitatively, a disguised sample has a similar feature representation with that of a copyrighted sample, i.e., $D_2(\mathcal{E}(x_c),\mathcal{E}(x_d))\leq\gamma_2$. As we have provided the feature threshold $\gamma_2$ which is sufficient for replicating the copyrighted content for different tasks, one can use this as a reference threshold to detect possible disguises. Specifically, given an encoder $\mathcal{E}$ (which is usually easily accessible), a copyrighted image $x_c$ which needs examination for infringement, one simply goes through the training set and computes $\mathcal{E}(x)$ for every single sample and compare with that of $x_c$. However, such a search alone only extracts suspects, which may not be true disguises. To rule out the sheer chance of collisions in the feature space, we perform another examination below.

(2) \textbf{Encoder-decoder examination:}
The encoder architecture in LDMs is part of an autoencoder architecture (e.g., the KL-based VAE), where the encoder and decoder are used separately for encoding and inference. Consequently, the decoder $\mathcal{D}$ can be naturally used to detect disguises qualitatively. Specifically, for a well-trained autoencoder, $\mathcal{D}(\mathcal{E}(x_c))\approx x_c$, while for disguises $\mathcal{E}(x_d)\approx \mathcal{E}(x_c)$, thus we have $\mathcal{D}(\mathcal{E}(x_d))\approx x_c$. In \Cref{fig:detection}, we show that the encoder-decoder architecture is a great detection tool for disguises, where the output of the autoencoder reveals the copyrighted content hidden in $x_d$.

\subsection{A broader definition of access}

\label{sec:acknowledgement}

Our experiments have demonstrated that by generating disguises, one can indirectly acquire access to copyrighted material as part of the training data to fuel LDM-based models, which could reproduce the copyrighted samples during inference or deployment. We expand the definition of \emph{direct access} to \emph{acknowledgment} to cover such scenarios as possible copyright infringement: any training data $x$ that contains similar latent information as that of copyrighted image $x_c$, measured by their similarity in the latent feature representation, even visually different from $x_c$, shall be considered to have an \emph{acknowledgment} of $x_c$.  The notion of \emph{acknowledgment} may also augment existing regulatory frameworks (e.g., the White House Executive Order on AI \cite{biden2023executive}) on AI governance in terms of data quality evaluation. Additionally, our quantification of \emph{acknowledgment} (the detection method) provide a timely tool for auditing beyond black-box access, which was pointed out to be insufficient in \cite{Casperetal24}.

\section{Conclusion}

In this paper, we review the current access criterion of containing copyrighted material in the training set (\emph{direct access}) in copyright infringement of generative models and point out its insufficiency by introducing disguised copyright infringement (\emph{indirect access}). Specifically, such an infringement is realized by injecting disguised samples into the training set, which urges LDMs to produce copyrighted content. Such disguises are generated with a simple algorithm and demonstrated to share the same concept with their target copyrighted images using the textual inversion tool. To alleviate the concern on the disguises, we expand the current visual auditing (browsing the training set) with additional tools, i.e., feature similarity search and encoder-decoder examination to better identify these disguises. Furthermore, we propose a broader definition of \emph{acknowledgment} to cover this new type of copyright violation.

\textbf{Limitations and future work:} One interesting future work is to quantify the number of disguises needed for reproducing in large-scale training, which can be further linked to the quantification of memorization of such models \cite{CarliniIJLTZ22, SimepalliSGGG23a,SimepalliSGGG23b,CarliniHNJSTBIW23,IppolitoTNZJLCC23,ZhangILJTC21}). Additionally, although our algorithms can generate descent disguises, we believe there is still room for improvement for optimization. Finally, one extension we didn't touch is the possibility of   ``chopping'' copyrighted data and hiding it in several images. It is intriguing to explore whether it is possible to generate such a smuggler's dataset and detection towards it.

\paragraph{Learning from noisy data} A simultaneous and independent work \citep{DarasDD24} considers learning LDMs from noisy data. Although the techniques are very different, our works share a similar implication: the training dataset may not
immediately resemble the generations produced, thus allowing copyright issues to be disguised from an auditor who manually inspects the dataset. In the work of \citet{DarasDD24}, the
training data is the original data with Gaussian noise applied to it.
In our case, the training data has the original data hidden in the
latent space.  In \Cref{sec:acknowledgement}, we argue that if training data contains the same latent information as some copyrighted data, then it should be acknowledged. \citet{DarasDD24} show that even when the latent space is corrupted the copyrighted information could still be contained, which could raise a stronger disguised copyright infringement attack.

\paragraph{Acknowledgment and fair use} Finally we discuss the anticipated impact of our definition of \emph{acknowledgment} on judicial decisions. Specifically, we believe \emph{acknowledgment} will not directly affect the current fair use doctrine or change its decision. To further illustrate the above claim, we first propose two principles: (1) copyright infringement is not fair use: although generative AI law is not rigorously defined, we generally assume that ``reproducing the copyrighted content'' is not fair use when there's direct access;
(2) disguised infringement does not (appear to) use the copyrighted data: there is no decision of fair use doctrine here as there is no apparent ``usage'' of the copyrighted data, not to mention ``fair use''. Thus the decision is on the ``originality'' of the output (reproduction of the copyrighted content) but not fair use. As a result, the notion of \emph{acknowledgment} aims to identify disguised copyright infringement in the second principle and challenge the decision of ``originality''. Based on the acquired outcome (reproduction), there would be no scenarios that \emph{acknowledgment} can change the decision of fair use according to the first principle.

\section*{Impact Statement}
In this paper, we adopt a simple algorithm to disguise images, which we realize may be applied as a concealed copyright infringement tool. We are surprised at how easy it is for Generative AI to circumvent the copyright law, which was originally designed for regulating human scrapers. This seemingly undetectable infringing threat may bring disadvantages for human artists, who are already losing the battle with GenAI. 
Furthermore, disguised images may not only be a concern to copyright owners but also a threat to Generative AI companies: if these disguises are accidentally collected as part of the training data, copyright infringement could be triggered unconsciously and unwillingly. 
As computer scientists, we are obligated to reveal the existence of such possible unlawful acts and provide technical tools to identify such behaviors. We expect our paper to be a prudent tool such that disguised copyright infringement and its corresponding detection methods are recognized by law experts. 

\section*{Acknowledgements}
We thank the reviewers for the critical comments that have largely improved the presentation and precision of this paper.
We gratefully acknowledge funding support from NSERC and the Canada CIFAR AI Chairs program. 
YY is also supported by the Ontario early researcher program. 
Resources used in preparing this research were provided, in part, by the Province of Ontario, the Government of Canada through CIFAR, and companies sponsoring the Vector Institute.

\printbibliography[segment=0,title={References}]

\newpage\clearpage
\appendix
\onecolumn
\section{Additional Experiments on Revealing Disguises}
\label{app:add_exp}

We provide additional experimental results on revealing disguises to further support our observations.

\paragraph{Textual inversion on $x_c$ and $x_b$:}

Recall that \Cref{fig:watermark}, \Cref{fig:objective} and \Cref{fig:style_main} show the disguised images $x_d$ and the new images generated by textual inversion. To further show the effect of our algorithm on generating disguises, we directly perform a textual inversion (with the same setting in our main experiments) on the copyrighted images $x_c$ and the base images $x_b$ below. Firstly, for copyrighted symbols:

Learning on the copyrighted images $x_c$:

\begin{figure*}[ht]
    \centering
    {\includegraphics[width=0.2\textwidth]{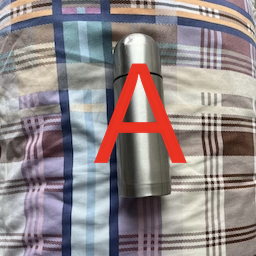}}
    \hspace{-0.18cm}
    {\includegraphics[width=0.2\textwidth]{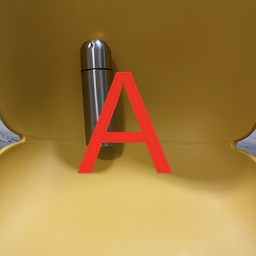}}
    \hspace{-0.18cm}
    {\includegraphics[width=0.2\textwidth]{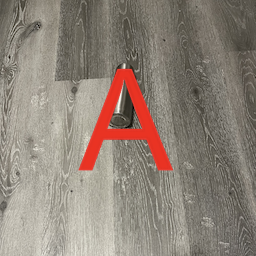}}
    \hspace{-0.18cm}
    {\includegraphics[width=0.2\textwidth]{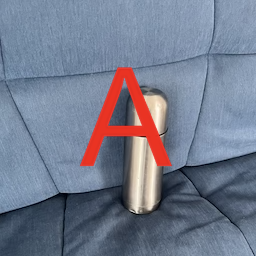}}
    {\includegraphics[width=0.8\textwidth]{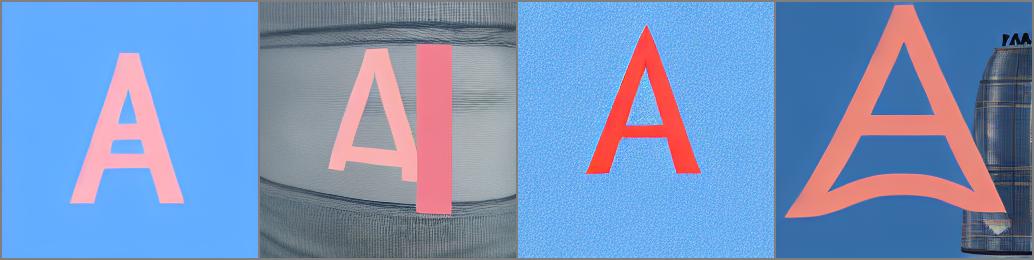}}

    \caption{Textual inversion on images $x_c$ with the copyrighted symbol ``A'' . The first row shows the copyrighted images $x_c$, and the second row shows new images generated by textual inversion after learning on the above $x_c$. }
    \label{fig:watermark_target}
\end{figure*}

Learning on the base images $x_b$:

\begin{figure*}[ht]
    \centering
    {\includegraphics[width=0.2\textwidth]{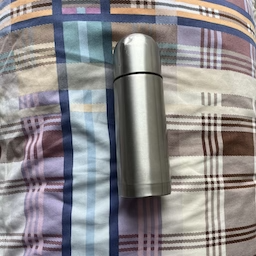}}
    \hspace{-0.18cm}
    {\includegraphics[width=0.2\textwidth]{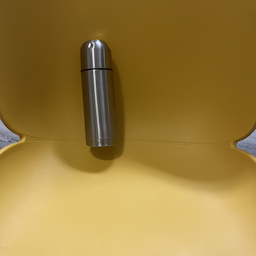}}
    \hspace{-0.18cm}
    {\includegraphics[width=0.2\textwidth]{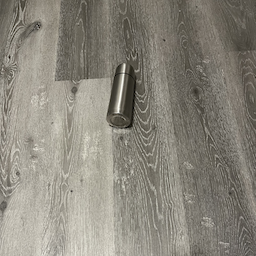}}
    \hspace{-0.18cm}
    {\includegraphics[width=0.2\textwidth]{images/watermark/bottle_clipped/4.png}}

    {\includegraphics[width=0.8\textwidth]{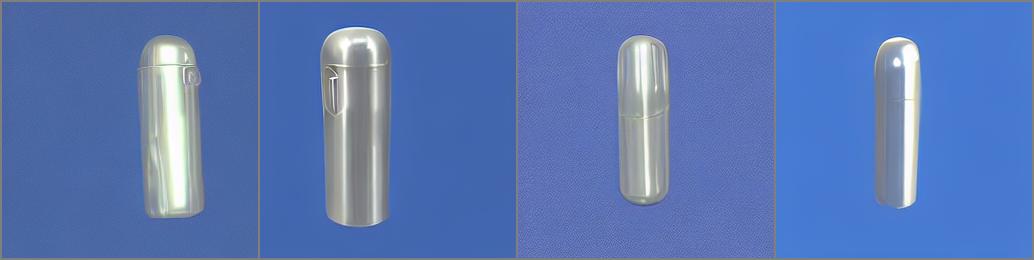}}

    \caption{Textual inversion on base images $x_b$ (different images of a bottle). The first row shows the base images $x_b$, and the second row shows new images generated by textual inversion after learning on the above $x_b$.}
    \label{fig:watermark_base}
\end{figure*}
Comparing with \Cref{fig:watermark}, we observe that although the disguises $x_d$ look visually similar to their corresponding base images $x_b$, they contain drastically different latent information.

\newpage
Next, we show the results for copyrighted content in the same format:

Learning on the copyrighted images $x_c$:

\begin{figure*}[ht]
    \centering  {\includegraphics[width=0.2\textwidth]{images/entire_image/sunflower_clipped/5.jpg}}
    {\includegraphics[width=0.2\textwidth]{images/entire_image/sunflower_clipped/6.jpg}}
    {\includegraphics[width=0.2\textwidth]{images/entire_image/sunflower_clipped/2.jpg}}
    
    {\includegraphics[width=0.8\textwidth]{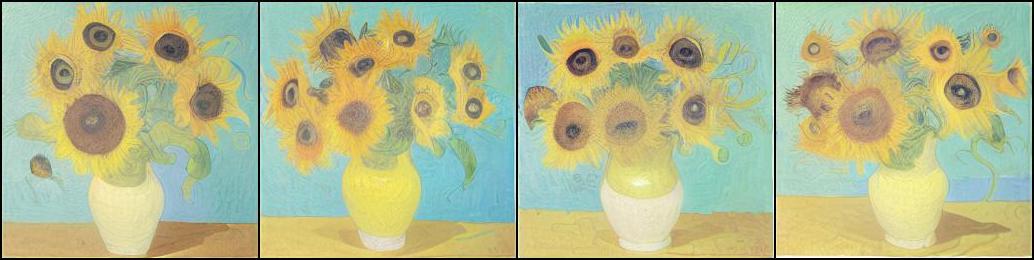}}

    \caption{Textual inversion on copyrighted images $x_c$ (\emph{The Sunflowers} by Vincent Van Gogh). The first row shows the copyrighted images $x_c$, and the second row shows new images generated by textual inversion after learning on the above $x_c$.}
    \label{fig:obj_target}
\end{figure*}

Learning on the base images $x_b$:

\begin{figure*}[ht]
    \centering
    {\includegraphics[width=0.2\textwidth]{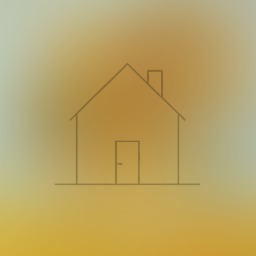}}
    {\includegraphics[width=0.2\textwidth]{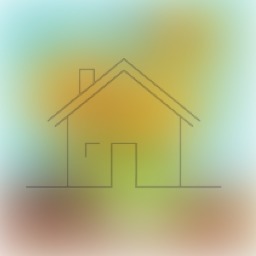}}
    {\includegraphics[width=0.2\textwidth]{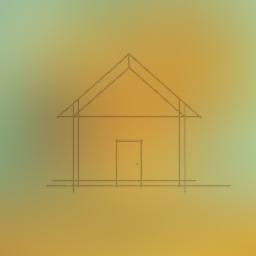}}

    {\includegraphics[width=0.8\textwidth]{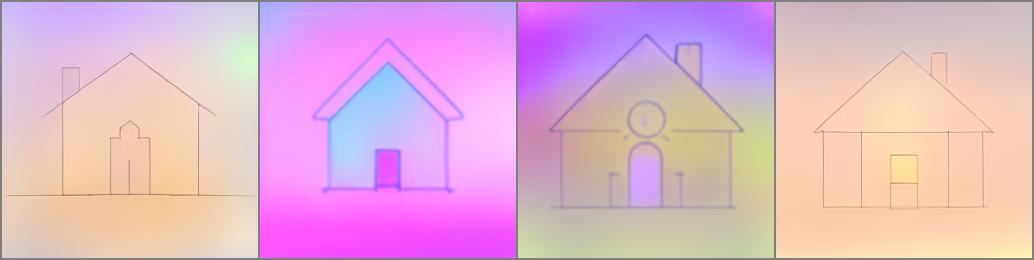}}
    
    \caption{Textual inversion on base images $x_b$ (images of sketched houses with the blurry version of the corresponding $x_c$ as the background). The first row shows the base images $x_b$, and the second row shows new images generated by textual inversion after learning on the above $x_b$.}
\end{figure*}

Compared to \Cref{fig:objective}, we observe that the background (blurry version of the $x_c$) does not provide any information regarding the copyrighted images $x_c$, and the success of the disguises is accomplished by our generating algorithm.

\newpage
Finally, we show the results for style scraping:

Learning on the copyrighted images $x_c$:

\begin{figure*}[ht]
    \centering
    {\includegraphics[width=0.2\textwidth]{images/style/Fig5/vg1.jpg}}
    {\includegraphics[width=0.2\textwidth]{images/style/Fig5/vg2.jpg}}
    {\includegraphics[width=0.2\textwidth]{images/style/Fig5/vg3.jpg}}
    {\includegraphics[width=0.8\textwidth]{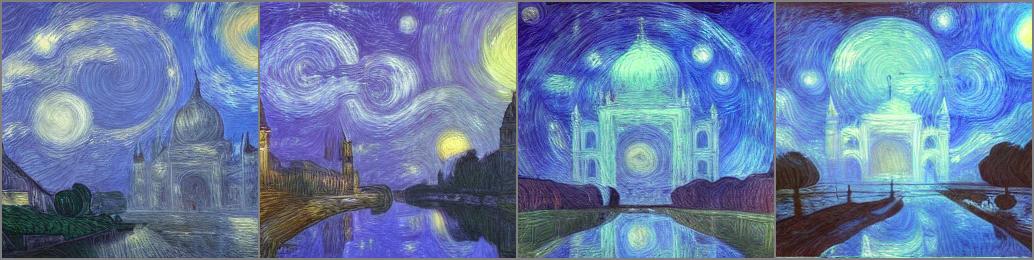}}

    \caption{Textual inversion on copyrighted images $x_c$ (images in the style of \emph{The Starry Night} by Vincent Van Gogh). The first row shows the copyrighted images $x_c$, and the second row shows new images generated by textual inversion after learning on the above $x_c$.}
\end{figure*}

Learning on the base images $x_b$:

\begin{figure*}[ht]
    \centering
{\includegraphics[width=0.2\textwidth]{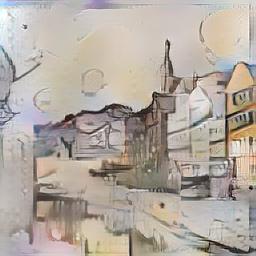}}
{\includegraphics[width=0.2\textwidth]{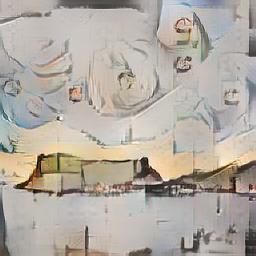}}
{\includegraphics[width=0.2\textwidth]{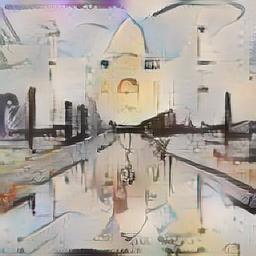}}

{\includegraphics[width=0.8\textwidth]{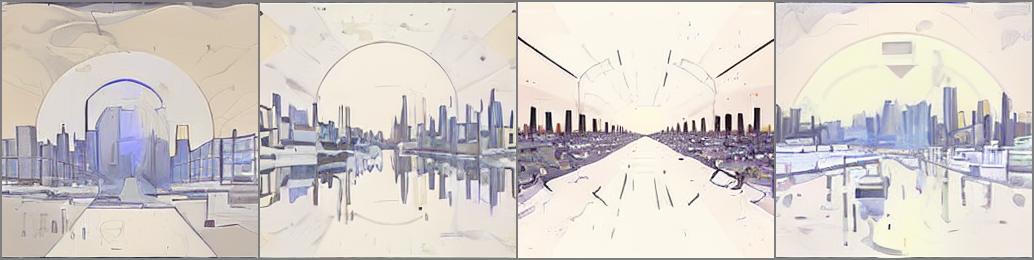}}

    \caption{Textual inversion on base images $x_b$ (images in watercolor style). The first row shows the base images $x_b$, and the second row shows new images generated by textual inversion after learning on the above $x_b$.}
\end{figure*}

Compared to \Cref{fig:style_main}, we again confirm that the disguises $x_d$ generated for style scraping look visually similar to their corresponding base images $x_b$ but contain latent information similar to $x_c$.

\begin{figure}[ht]
    \centering
{\includegraphics[width=0.2\textwidth]{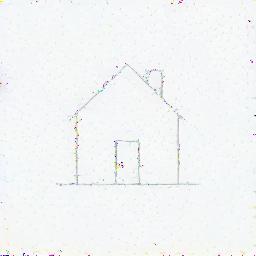}}
{\includegraphics[width=0.2\textwidth]{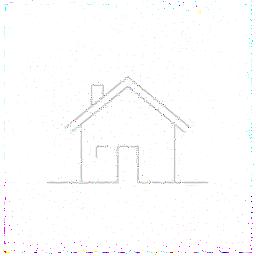}}
{\includegraphics[width=0.2\textwidth]{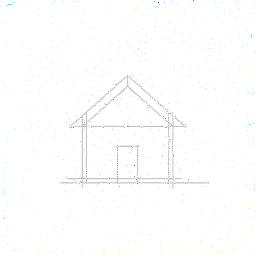}}

{\includegraphics[width=0.6\textwidth]{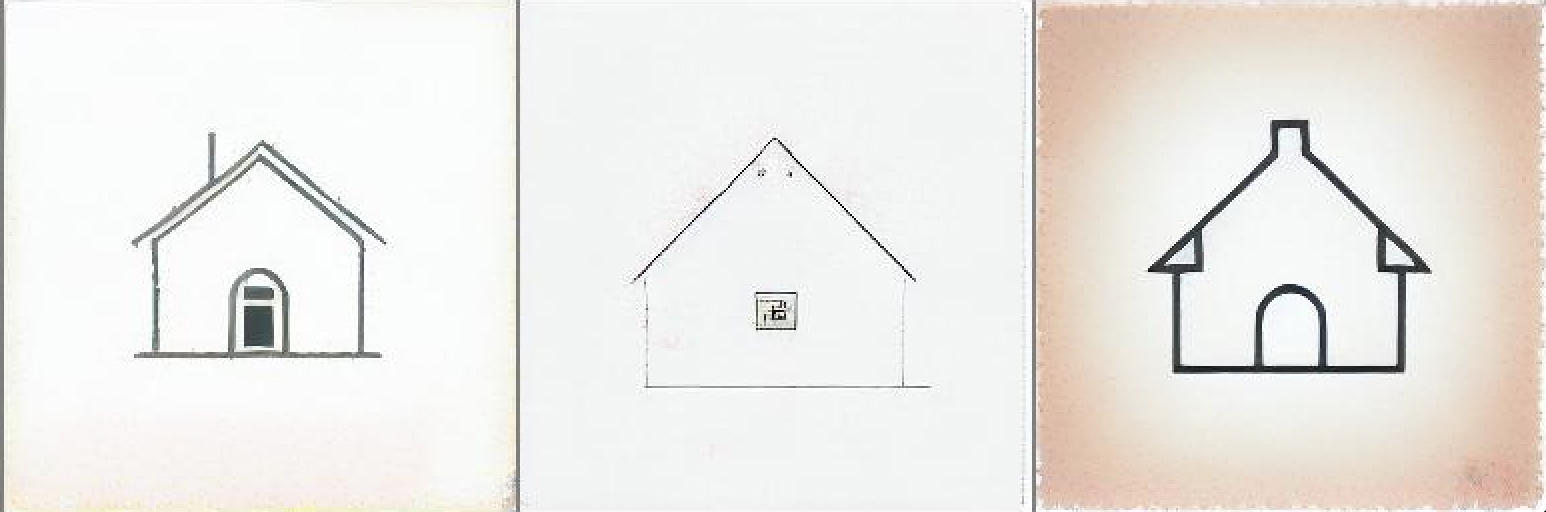}}

    \caption{Failure to learn the copyrighted content (\emph{The Sunflowers}) from a disguised sketch of the house with a white background. The first row:  disguises; the second row: images generated by textual inversion.}
    \label{fig:fail}
\end{figure}

\paragraph{Choice of $x_b$ for generating disguised content:} Recall that for generating disguises in \Cref{fig:objective},  we choose a blurry version of $x_c$ as background to retain the color pattern. Here we demonstrate the necessity of the background by substituting it for a white background as base images $x_b$. In \Cref{fig:fail}, we observe that the disguises fail in reproducing the copyrighted content on textual inversion.

\begin{figure*}[hb]
    \centering
    \subfloat[]{\includegraphics[width=0.19\textwidth]{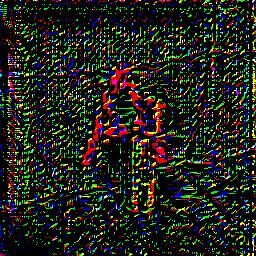}} \hfill
    \subfloat[]{\includegraphics[width=0.76\textwidth]{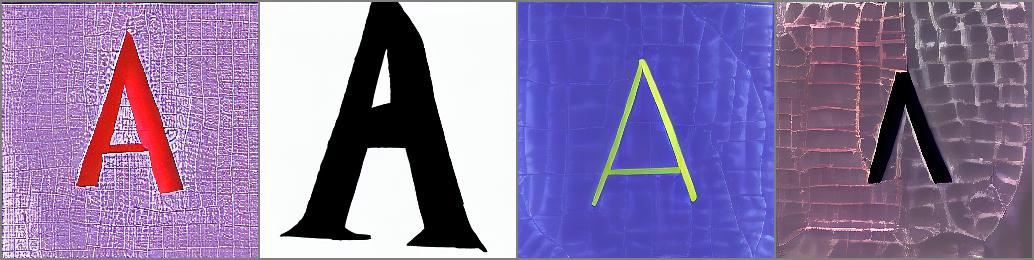}}
        \vspace{-2em}
        \subfloat[Disguises $x_d$]{\includegraphics[width=0.19\textwidth]{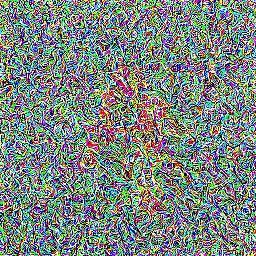}} \hfill
    \subfloat[Inference on textual inversion by learning on $x_d$ only ]{\includegraphics[width=0.76\textwidth]{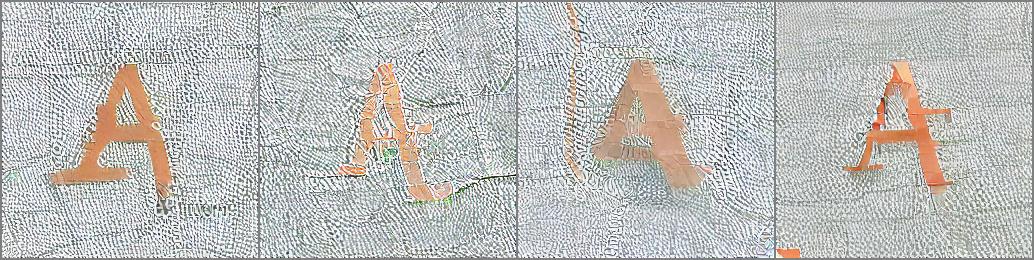}}

    \caption{Textual inversion on the disguised symbol without any input constraints. We deploy different initializations for generating disguises: the first row: zero tensor; the second row: Gaussian noise. }
    \label{fig:init}
\end{figure*}

\newpage
\paragraph{Intialization for unbounded disguises:} In \Cref{fig:noise_watermarking} (first column), we show the disguises generated without any input constraints. Note that the disguise (for the copyrighted symbol) still contains the base image as background as the initialization of the disguise is $x_b$. To confirm, we show two different initializations (zero tensor and Gaussian noise) in \Cref{fig:init} and observe that the background diminishes while the disguises still contain the copyrighted symbol.

\section{Data augmentation}
\begin{figure*}[hb]
    \centering
    \subfloat[Disguise $x_d$]{\includegraphics[width=0.15\textwidth]{images/watermark/bottle_clipped/poisoned3.jpg}}
    \subfloat[Learn on $x_d$]{\includegraphics[width=0.15\textwidth]{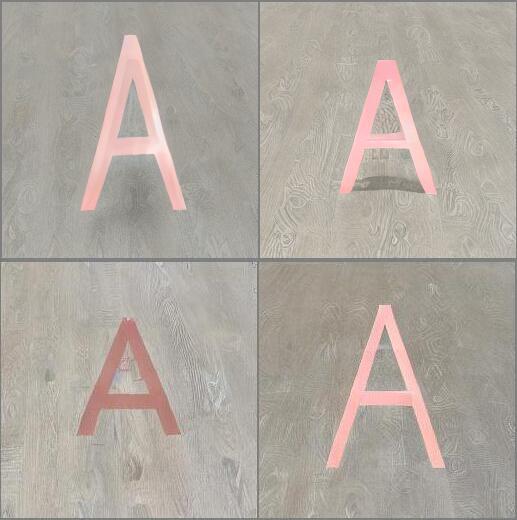}}
    \subfloat[Learn on $\textit{hflip}(x_d)$]{\includegraphics[width=0.15\textwidth]{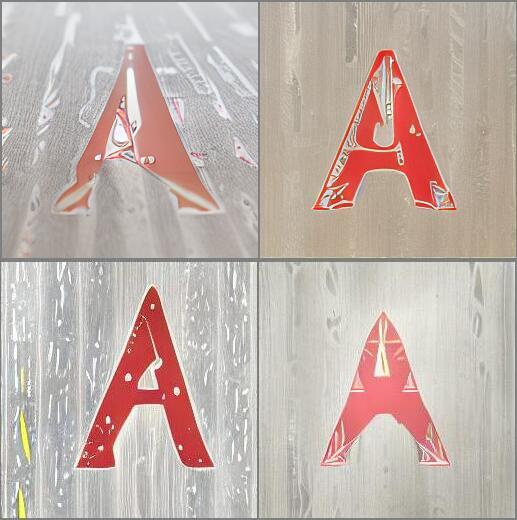}}
    \hfill
    \subfloat[Disguise $x_d$]{\includegraphics[width=0.15\textwidth]{images/entire_image/sunflower_clipped/poisoned2.jpg}}
    \subfloat[Learn on $x_d$]{\includegraphics[width=0.15\textwidth]{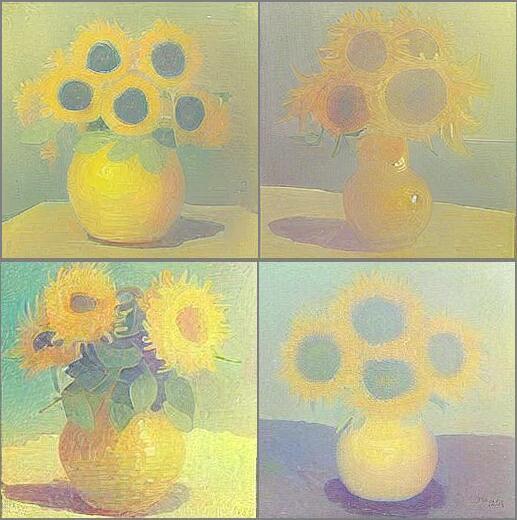}}
    \subfloat[Learn on $\textit{hflip}(x_d)$]{\includegraphics[width=0.15\textwidth]{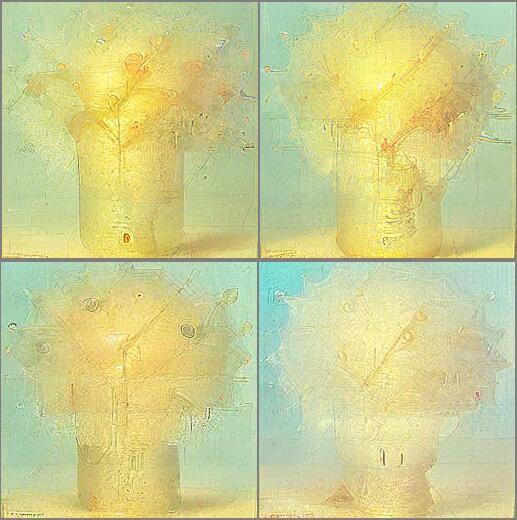}}
    \\
    \subfloat[Disguise $x_d$]{\includegraphics[width=0.15\textwidth]{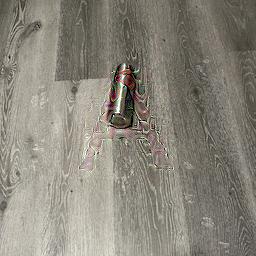}}
    \subfloat[Learn on $x_d$]{\includegraphics[width=0.15\textwidth]{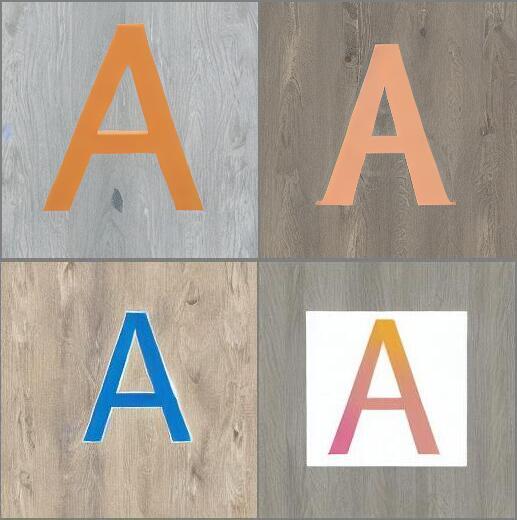}}
    \subfloat[Learn on $\textit{hflip}(x_d)$]{\includegraphics[width=0.15\textwidth]{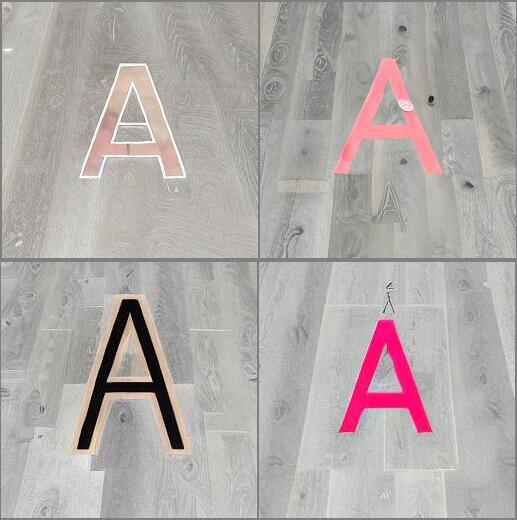}}
    \hfill
    \subfloat[Disguise $x_d$]{\includegraphics[width=0.15\textwidth]{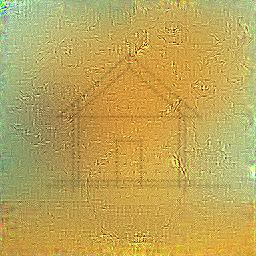}}
    \subfloat[Learn on $x_d$]{\includegraphics[width=0.15\textwidth]{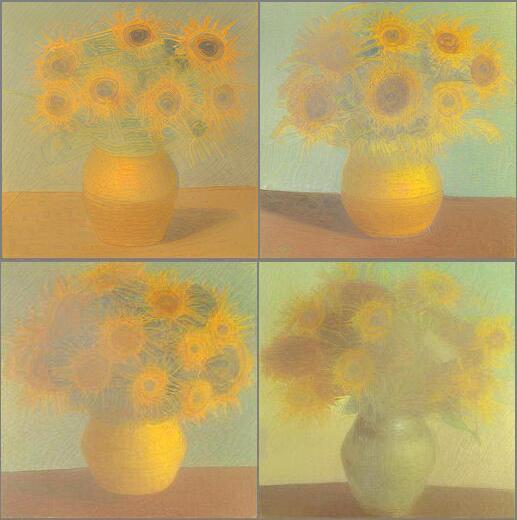}}
    \subfloat[Learn on $\textit{hflip}(x_d)$]{\includegraphics[width=0.15\textwidth]{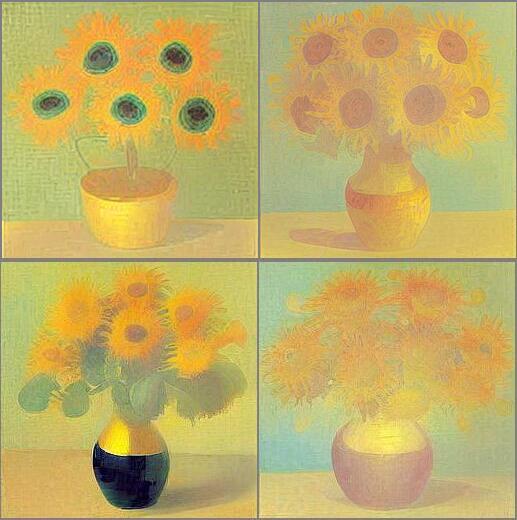}}

    \caption{Textual inversion on different disguises. Top left: textual inversion on a single watermarking disguise. Top right: textual inversion on a single content disguise. Bottom row: same as top row, but with robust disguises.}
    \label{fig:aug}
\end{figure*}

In textual inversion, images are horizontally flipped with 50\% probability. In this section, we explore how this affects the success of our disguises. The top two rows of \Cref{fig:aug} show the results of training textual inversion on our disguise as well as its horizontally flipped augmentation. For the simpler concept of watermarking, textual inversion learns a concept similar to the intended one even when the input image is flipped. On the other hand, when the concept becomes more difficult, textual inversion struggles to learn the concept under a horizontal flip augmentation (top right of \Cref{fig:aug}).

To ensure that a horizontal flip augmentation would not damage the quality of our poison, we construct a new robust poison that additionally penalizes the distance between the features of the horizontally flipped poison and the horizontally flipped copyrighted image $x_c$. This can be done with a simple change to the objective from
\begin{align}
\argmin_{x_d} ~\alpha D_1(x_b, x_d) + D_2(\mathcal{E}(x_c), \mathcal{E}(x_d))
\end{align}
to
\begin{align}
\argmin_{x_d} ~\alpha D_1(x_b, x_d) + D_2(\mathcal{E}(x_c), \mathcal{E}(x_d)) + D_2(\mathcal{E}(\textit{hflip}(x_c)), \mathcal{E}(\textit{hflip}(x_d))).
\end{align}
The bottom two rows of \Cref{fig:aug} demonstrate the effectiveness of the robust poison and its qualitative improvement under horizontal flipping.

\begin{table*}[hbt!]
    \centering
    \caption{The encoder-decoder examination's threshold $\zeta$, mean construction losses, false positive rate, and binary AUROC across symbol, content, and style disguises. The second group of rows is for the setting of mixture with base images, and the third group of rows is for the setting of mixture with 100 randomly selected Imagenette images.}
    \label{table:detection}
    \setlength\tabcolsep{6pt}
   \resizebox{0.48\textwidth}{!}{\begin{tabular}{lrrrr}

 \toprule
  & Symbol & Content & Style \\
 \midrule
  $\mathbb{E}_{x_d}(\mathcal{L}_{\text{reconstruct}}(x_d))$ & 0.0856 & 0.1632 & 0.2667 \\
 Threshold $\zeta$& 0.0703 & 0.1412 & 0.2506 \\
 \midrule
  $\mathbb{E}_{x_b}(\mathcal{L}_{\text{reconstruct}}(x_b))$ & 0.0502 & 0.0256 & 0.0551 \\
 FPR ($x_b$ misclassified) & 1/4 & 0/3 & 0/3 \\
 \bf AUC & 0.8750 & 1 & 1 \\
\midrule
  $\mathbb{E}_{x_\text{clean}}(\mathcal{L}_{\text{reconstruct}}(x_\text{clean}))$ & 0.0655 & - & - \\
 FPR ($x_\text{clean}$ misclassified) & 42/100 & 2/100 & 0/100 \\
 \bf AUC & 0.7550 & 0.9933 & 1 \\
 \bottomrule

\end{tabular}}
\vspace{-10pt}
\end{table*}

\section{Quantitative detection}

In \Cref{sec:detection}, we show how to qualitatively distinguish a disguise with encoder-decoder examination. In this section, we further examine two quantitative detection scenarios with/without the presence of the copyrighted image $x_c$:

\textbf{Without the presence of $x_c$:} Specifically, we observe that in \Cref{fig:detection}, for disguises $x_d$, the input $x_d$ and output of the autoencoder $\mathcal{D}(\mathcal{E}(x_d))$ are significantly different, i.e., the reconstruction loss $D_1(\mathcal{D}(\mathcal{E}(x_d)), x_d)$ is ``big'' (recall that $D_1$ is the distance measure in the input space, a sum of MS-SSIM loss and $L_1$ loss ). This property differs from normal images, which are expected to have low reconstruction loss. Leveraging this intuition, we develop a detection criterion for disguised images:
\begin{align}
x \text{ is a disguise if } \mathcal{L}_{\text{reconstruct}}(x) := D_1(\mathcal{D}(\mathcal{E}(x)), x) \geq \zeta \label{eq:detect}
\end{align}
for some threshold $\zeta$. The choice of $\zeta$ differs on different tasks (depending on the choice of the task, which we specify below).
Note that this detection mechanism does not require any knowledge of the copyrighted image and is simply a data sanitization method that can be applied to any dataset.


In \Cref{table:detection}, we show the effectiveness of our proposed method across the three tasks we performed in \Cref{sec:exp} (copyrighted symbol, content, and style). \emph{Row 2} shows the mean of the reconstruction loss for the disguises we generated for each task; \emph{Row 3} shows the choice of threshold $\zeta$ as the lowest reconstruction loss out of all the disguised images such that there are no false negatives. \emph{Rows 4-6} examine the detection performance in terms of FPR (False Positive Rate) and AUC (Are Under the Curve) within the pool of the mix of the disguises $x_d$ and their corresponding base images $x_b$; \emph{Rows 7-9} performs the same examination on a larger pool of the mix of the disguises $x_d$ and a subset of 100 images randomly chosen from ImageNet.

We observe that for copyrighted symbols, as the copyrighted images $x_c$ (see \Cref{fig:watermark_target}) and the base images $x_b$ (see \Cref{fig:watermark_base}) share the same background and only differ on the symbol, the reconstruction loss of $x_d$ is low (mean of 0.0856) and falls into the range of that of clean images, thus having a higher FPR and a lower AUC than the other two tasks.

\textbf{With the presence of $x_c$:} In practical scenarios (e.g., examination in court), one may already have acquired the copyrighted images $x_c$ and want to find its (possible) corresponding disguise from a dataset. In such scenarios, one may directly use the detection method introduced in \Cref{sec:detection}. We repeat the above experiment using our feature similarity search as a first step, which we recall, screen disguises using the criterion $D_2(\mathcal{E}(x_c),\mathcal{E}(x_d))\leq\gamma_2$. We acquire perfect AUC score and obtain no false positives across all tasks, which indicate that the second step (encoder-decoder examination) is not required for this specific task.

\begin{figure*}[hbt!]
    \centering

    \subfloat[]{\includegraphics[width=0.2\textwidth]{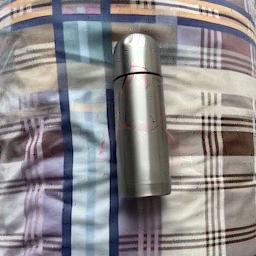}}
    \subfloat[]{\includegraphics[width=0.2\textwidth]{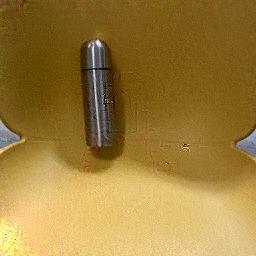}}
    \subfloat[]{\includegraphics[width=0.2\textwidth]{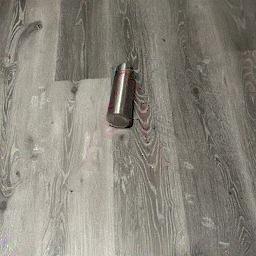}}
    \subfloat[]{\includegraphics[width=0.2\textwidth]{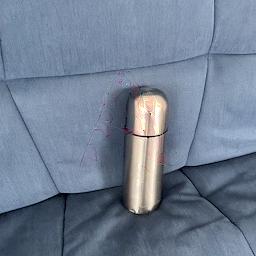}}
    \vspace{-2em}
    \subfloat[]{\includegraphics[width=0.2\textwidth]{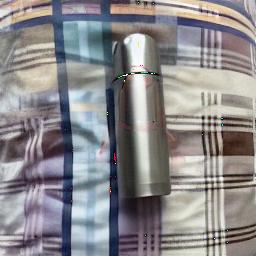}}
    \subfloat[]{\includegraphics[width=0.2\textwidth]{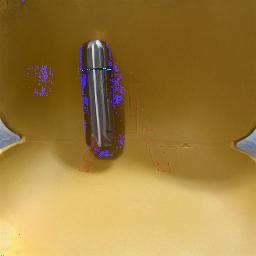}}
    \subfloat[]{\includegraphics[width=0.2\textwidth]{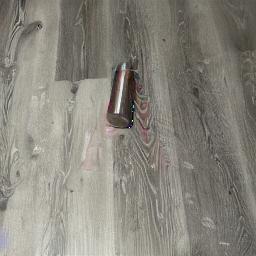}}
    \subfloat[]{\includegraphics[width=0.2\textwidth]{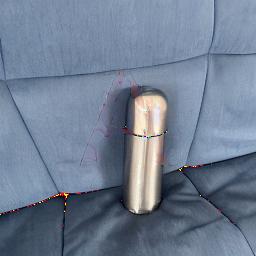}}
    \vspace{-2em}
    \subfloat[]{\includegraphics[width=0.8\textwidth]{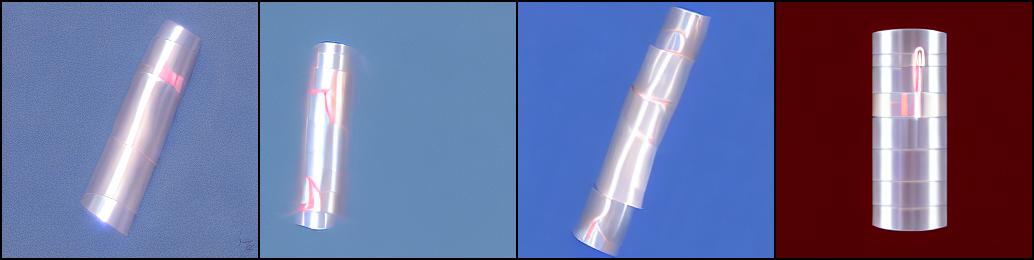}}

    \caption{An attempt to evade the encoder-decoder detection on symbols. The first row shows the disguises $x_d$, and the second row includes the output of the autoencoder $\mathcal{D}(\mathcal{E}(x_d))$ that is unable to reveal the copyrighted information contained in the disguises. The third row shows the (unsuccessful) outcome of applying textual inversion to learn the hidden symbol.}
    \label{fig:detection-evasion}
\end{figure*}

\section{Circumventing the detection?}

A stronger disguise may be generated to circumvent our detection method (e.g., the encoder-decoder examination).
This can be possibly accomplished by modifying the disguise generation objection. Recall that our previous  objective is:
\begin{align}
    \argmin_{x_d}\alpha D_1(x_b, x_d) + D_2(\mathcal{E}(x_c), \mathcal{E}(x_d)),
\end{align}
Here we wish to also minimize the input distance between $x_d$ and its autoencoder output  $D_1(\mathcal{D}(\mathcal{E}(x_d)), x_d))$, thus we can modify our objective to:

$$\argmin_{x_d}\alpha (D_1(x_b, x_d) + D_1(\mathcal{D}(\mathcal{E}(x_d)), x_d)) + D_2(\mathcal{E}(x_c), \mathcal{E}(x_d))$$

We perform experiments on the copyrighted symbols and observe in \Cref{fig:detection-evasion} that, the additional term adds a constraint such that $x_d$ and $\mathcal{D}(\mathcal{E}(x_d))$ are visually similar (row 2). However, performing textual inversion on these images also fails to reveal the copyrighted symbol. We conclude that this method might not be ideal, and additional study might be required.

\section{Additional experiments on disguised style}

In this section, we exhibit additional experimental results on style transfer by extending the choices of the (designated) copyrighted style $x_c$ and base images $x_b$. 

\begin{figure*}[ht!]
    \centering
    \subfloat[(1) \emph{San Giorgio Maggiore at Dusk}]{{\includegraphics[width=0.25\textwidth]{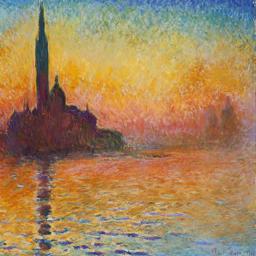}}}
    \subfloat[(2) \emph{The Starry Night}]{{\includegraphics[width=0.25\textwidth]{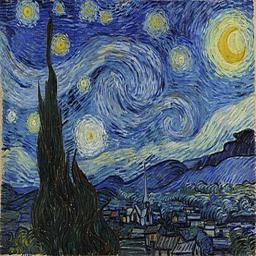}}}
    \subfloat[$x_c$: (1) with the style of (2)]{{\includegraphics[width=0.25\textwidth]{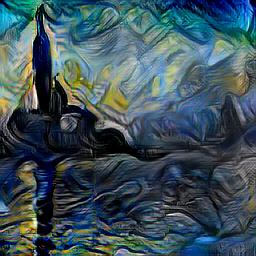}}}

    \caption{Creating the copyrighted image $x_c$ (third column) using AdaIN-based style transfer \cite{HuangB17}. $x_c$ is \emph{San Giorgio Maggiore at Dusk} by Claude Monet transferring to the style of \emph{The Starry Night} by Vincent Van Gogh. }
    \label{fig:monet}
    \vspace{-10pt}
\end{figure*}

\begin{figure*}[hbt!]
\vspace{-10pt}
    \centering
    \subfloat[{Base $x_b$}]{{\includegraphics[width=0.2\textwidth]{images/style/Monet/monet.jpg}}}
    \subfloat[{Disguise $x_d$}]{{\includegraphics[width=0.2\textwidth]{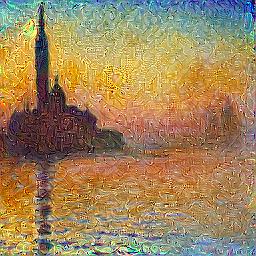}}}
    \subfloat[{$x_c$}]{{\includegraphics[width=0.2\textwidth]{images/style/Monet/target.jpg}}}\\
    \vspace{-10pt}
    \subfloat[Images generated by textual inversion after training on the $x_d$]{\includegraphics[width=0.8\textwidth]{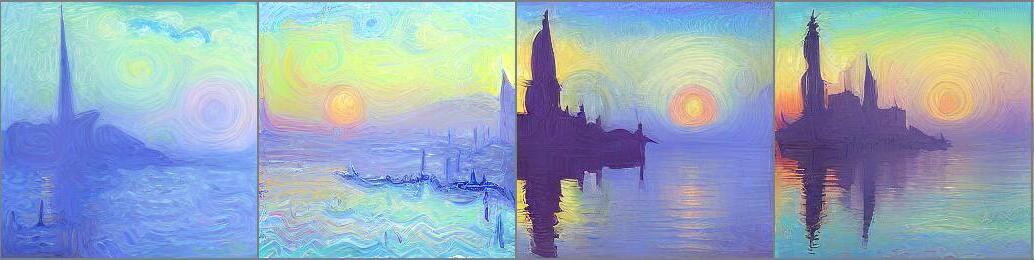}}

    \caption{Additional results on the disguised copyrighted style with textual inversion, where we generate disguise $x_d$ that visually resembles the style of Claude Monet but containing the style of Vincent Van Gogh in the latent space.}
    \label{fig:monet_2}
\end{figure*}

\textbf{Disguises in the style of Claude Monet}: Recall that in \Cref{sec:exp}, we choose the base images $x_b$ as the ones with the watercolor style. In the following experiments, we aim to change $x_b$ to images with the style of Claude Monet and still generate disguises $x_d$ that implicitly contain the style of Vincent Van Vogh. In \Cref{fig:monet}, we first create the $x_c$ using style transfer \cite{HuangB17}, which is \emph{San Giorgio Maggiore at Dusk} (1908-1912) by Claude Monet transferring to the style of \emph{The Starry Night} by Vincent Van Gogh. In \Cref{fig:monet_2}, we create the disguise $x_d$ (row 1, column 2) with the visual appearance of the base $x_b$ and show that the images generated by textual inversion after training on $x_d$ still contain the style of \emph{The Starry Night}.

\begin{figure*}[ht!]
    \vskip -10pt
    \centering
    \subfloat[(1) Taj Mahal \emph{(watercolor)}]{{\includegraphics[width=0.25\textwidth]{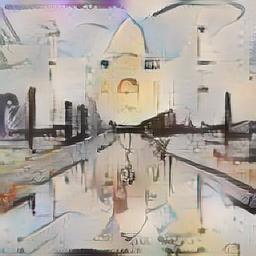}}}\hskip 0.3cm
    \subfloat[(2) \emph{Cecily Brown-56542}]{{\includegraphics[width=0.25\textwidth]{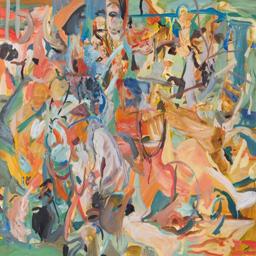}}}\hskip 0.3cm
    \subfloat[$x_c$: (1) with the style of (2)]{{\includegraphics[width=0.25\textwidth]{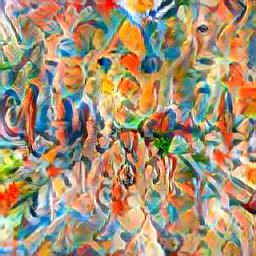}}}

    \caption{Creating the copyrighted image $x_c$ (third column) using AdaIN-based style transfer \cite{HuangB17}. $x_c$ is Taj Mahal transferring to the style of Cecily Brown.}
    \label{fig:cecily}
    \vspace{-10pt}
\end{figure*}

\begin{figure*}[ht]
\vskip -10pt
    \centering
    \subfloat[{Base $x_b$}]
    {\includegraphics[width=0.2\textwidth]{images/style/Cecily/water_building.jpg}}
    \subfloat[{Disguise $x_d$}]
    {\includegraphics[width=0.2\textwidth]{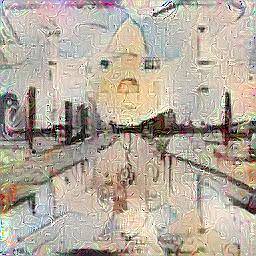}}
    \subfloat[{$x_c$}]
    {\includegraphics[width=0.2\textwidth]{images/style/Cecily/cecily_building.jpg}}
    
    \subfloat[Images generated by textual inversion after training on the $x_d$]
    {\includegraphics[width=0.8\textwidth]{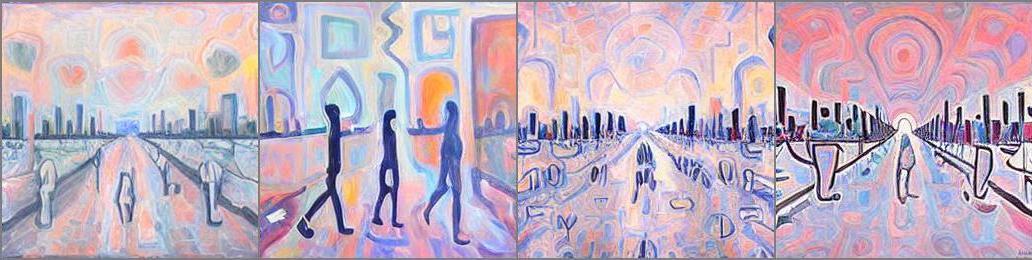}}
    \caption{Additional results on the disguised copyrighted style with textual inversion, where we generate disguise $x_d$ that visually resembles the watercolor style but containing the style of Cecily Brown in the latent space.}
    \label{fig:ceicly2}
\end{figure*}

\textbf{Scraping the style of other artists}: In the previous experiments, we choose the style of \emph{The Starry Night} as the designated copyrighted style. Next, we show the results for disguised style scraping of another artist Cecily Brown\footnote{\url{https://en.wikipedia.org/wiki/Cecily_Brown}}. In \Cref{fig:cecily}, we first create the $x_c$ using style transfer \cite{HuangB17}, which is Taj Mahal (adamkaz/Getty Images) transferring to the style of the painting with serial number \emph{56542} by Cecily Brown\footnote{\url{https://www.artnews.com/art-in-america/features/cecily-brown-56542/}}. In \Cref{fig:ceicly2}, we create the disguise $x_d$ (row 1, column 2) with the visual appearance of the base $x_b$ (row 1, column 1) and show that the images generated by textual inversion after training on $x_d$ contain the style of the style of Cecily Brown. 

\begin{figure*}[t!]
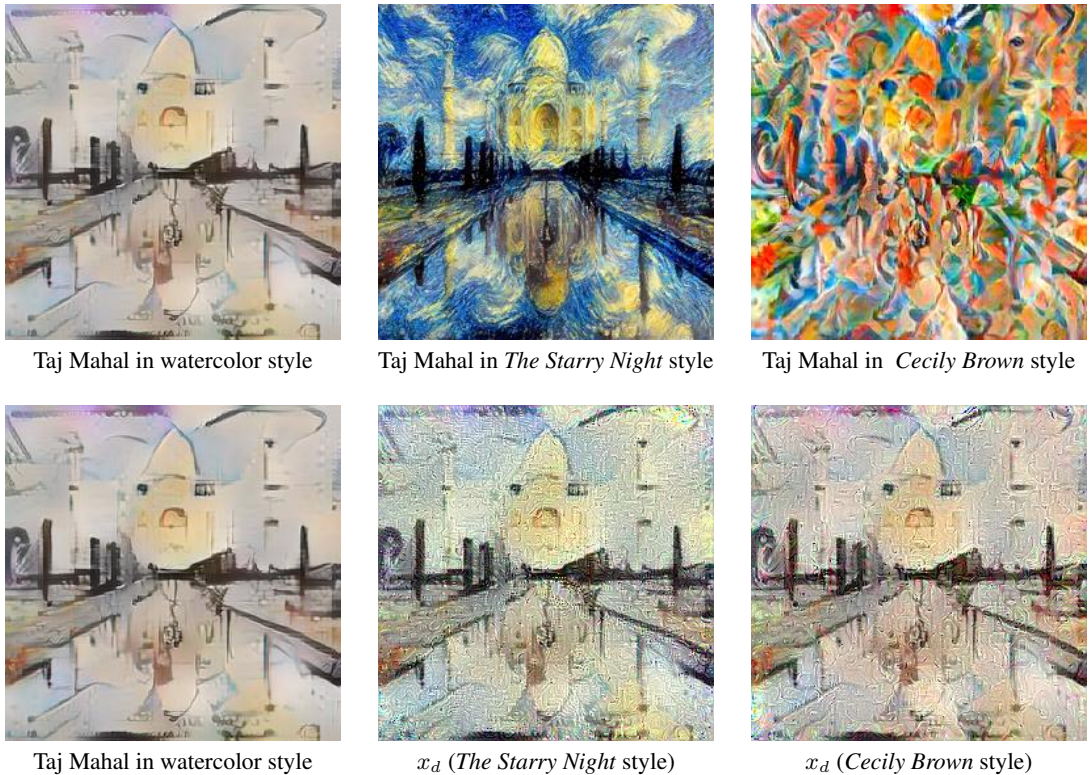

    \centering
    \subfloat[{Taj Mahal in watercolor style}]
    {\includegraphics[width=0.26\textwidth]{images/style/Appen/water_building.jpg}} \hskip 0.5cm
    \subfloat[Taj Mahal in \emph{The Starry Night} style]
    {\includegraphics[width=0.26\textwidth]{images/style/Fig5/vg3.jpg}}
    \hskip 0.5cm
    \subfloat[Taj Mahal in \emph{ Cecily Brown} style]
    {\includegraphics[width=0.26\textwidth]{images/style/Cecily/cecily_building.jpg}}

    \subfloat[{Taj Mahal in watercolor style}]
    {\includegraphics[width=0.26\textwidth]{images/style/Appen/water_building.jpg}}\hskip 0.5cm
    \subfloat[$x_d$ (\emph{The Starry Night} style)]
    {\includegraphics[width=0.26\textwidth]{images/style/Fig5/updated/poison_100000_building.jpg}}\hskip 0.5cm
    \subfloat[$x_d$ (\emph{Cecily Brown} style)]
    {\includegraphics[width=0.26\textwidth]{images/style/Cecily/poison.jpg}}

    \caption{Comparison between different disguises $x_d$ we generated that contain drastically different styles in the feature space, but are visually similar in the input space. }
    \label{fig:base}
    \vskip -0.5cm
\end{figure*}

Finally, in \Cref{fig:base}, we compare two different disguises $x_d$ and observe images that visually look similar can contain drastically different styles in the feature space, which again demonstrates the threat of disguised copyright infringement.

\begin{figure*}[htb!]
    \centering
    \subfloat[{Base $x_b$}]
    {\includegraphics[width=0.2\textwidth]{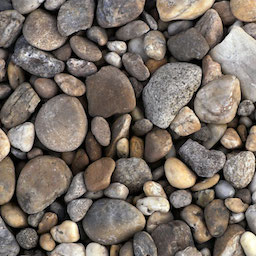}}
    \subfloat[{Disguise $x_d$}]
    {\includegraphics[width=0.2\textwidth]{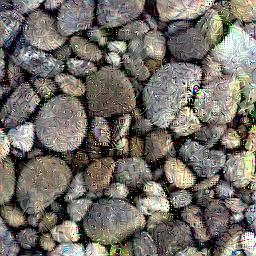}}
    \subfloat[{$x_c$}]
    {\includegraphics[width=0.2\textwidth]{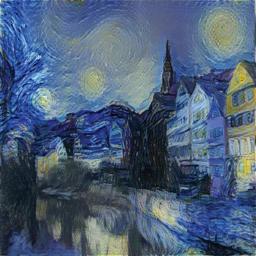}}
    \subfloat[{$\mathcal{D}(\mathcal{E}(x_d$))}]
    {\includegraphics[width=0.2\textwidth]{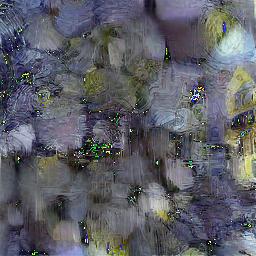}} \\
    \vskip -0.05cm
    \subfloat[Images generated by textual inversion after training on the $x_d$]
    {\includegraphics[width=0.8\textwidth]{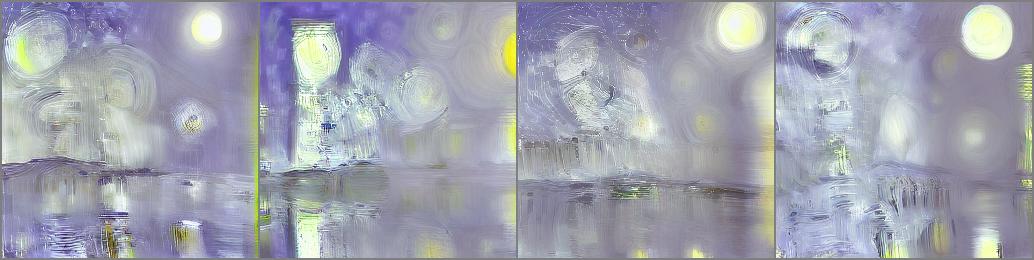}}

    \caption{Additional results on the disguised copyrighted style with textual inversion, where we generate disguise $x_d$ that visually resembles a random base image but intend to contain the style of \emph{The Starry Night} in the latent space. }
    \label{fig:random_base}
\end{figure*}

\textbf{Choice of $x_b$: } In previous experiments, we construct the disguises $x_d$ by choosing the base image $x_b$ to have a different style from $x_c$ but containing the same content. In \Cref{fig:random_base}, we also explore the possibility of choosing a random base image from ImageNet (row 1, column 1). Our results show that although the style is partially learned by textual inversion from the disguise $x_d$, it also loses the structural information of $x_c$.

\section{Disguises in the Wild}
\label{sec:dream_finetune}

We have previously \emph{revealed} the concept hidden in the disguised data using textual inversion, which is also an LDM-based application that can possibly generate copyright materials. In this section, we further extend our evaluation to other LDM-based pipelines that change the parameters of the U-Net. Specifically, we consider DreamBooth \citep{ruiz2023dreambooth}, which fine-tunes a pre-trained LDM model with only a few disguised images (similar to textual inversion); and mixed-training that contains both clean data and a small portion of disguises. 

\begin{figure}[t]
    \centering
    {\includegraphics[width=0.15\textwidth]{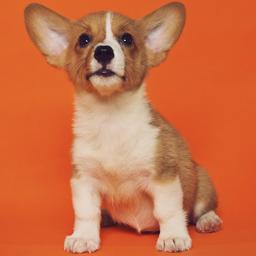}} 
    {\includegraphics[width=0.15\textwidth]{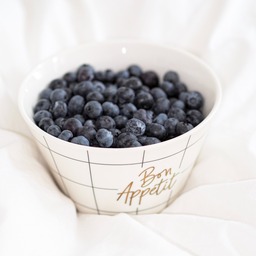}}
    \subfloat[$x_c$]  {\includegraphics[width=0.15\textwidth]{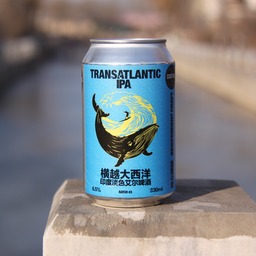}}
    {\includegraphics[width=0.15\textwidth]{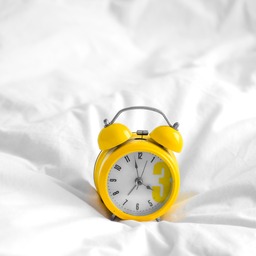}}
    {\includegraphics[width=0.15\textwidth]{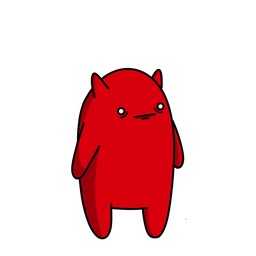}}
    \\
    \vskip -0.3cm
    {\includegraphics[width=0.15\textwidth]{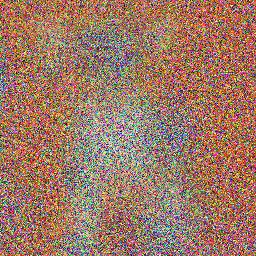}}
    {\includegraphics[width=0.15\textwidth]{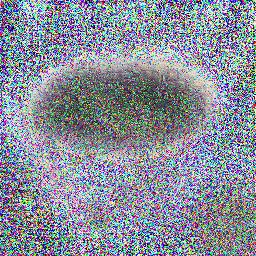}}
    \subfloat[Disguise $x_d$] {\includegraphics[width=0.15\textwidth]{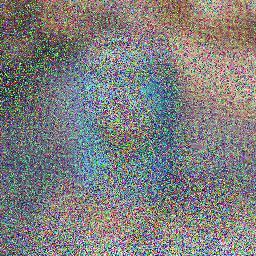}}
    {\includegraphics[width=0.15\textwidth]{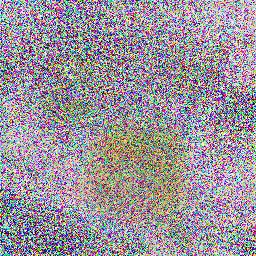}}
    {\includegraphics[width=0.15\textwidth]{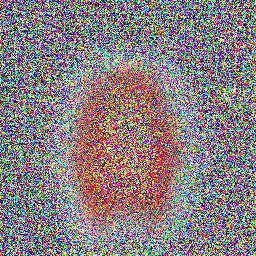}}
    \\
    \vskip 0.1cm
     \subfloat[Images generated by DreamBooth by fine-tuning with the above disguises]{{\includegraphics[width=0.15\textwidth]{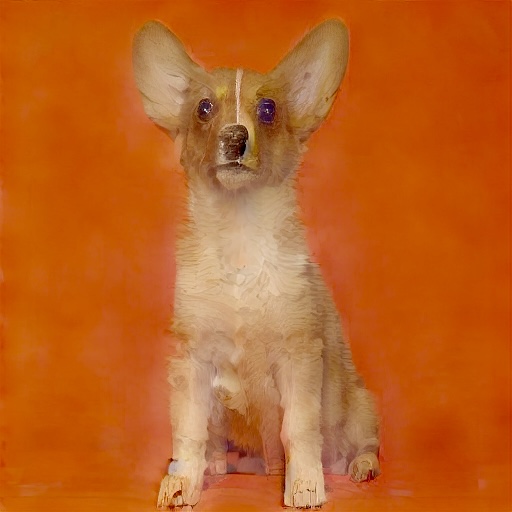}}
    {\includegraphics[width=0.15\textwidth]{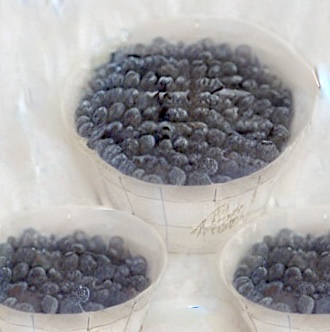}}
    {\includegraphics[width=0.15\textwidth]{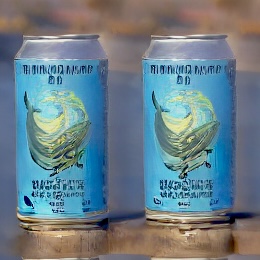}}
    {\includegraphics[width=0.15\textwidth]{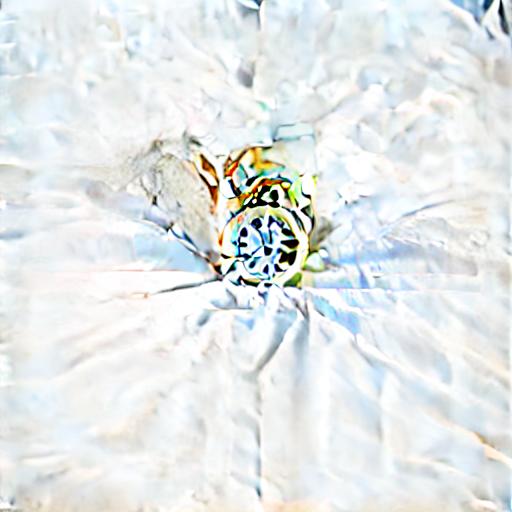}}
    {\includegraphics[width=0.15\textwidth]{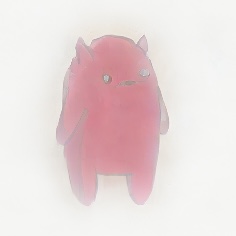}}}\\

    \caption{We show the disguised copyright infringement with DreamBooth. The first row: the designated copyrighted image $x_c$ from the DreamBooth dataset; the second row: the corresponding disguises $x_d$ generated with our \Cref{alg:DG}; the third row: images generated by DreamBooth after training on the above disguises $x_d$.}
    \label{fig:dreambooth}
\end{figure}

\begin{figure}[hbt!]
    \centering
    \subfloat[Images generated by LDM \emph{before} fine-tuning with our disguises using DreamBooth]{{\includegraphics[width=0.15\textwidth]{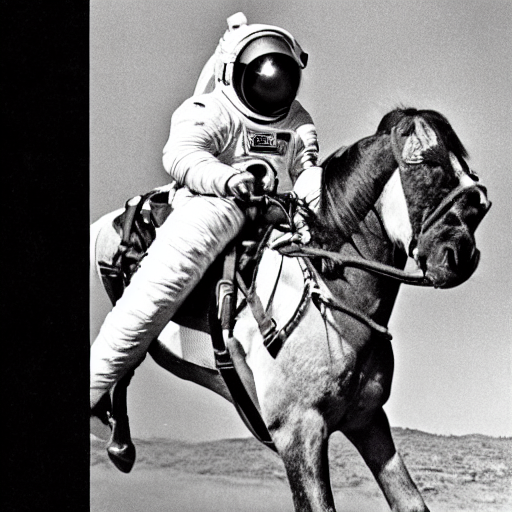}} 
    {\includegraphics[width=0.15\textwidth]{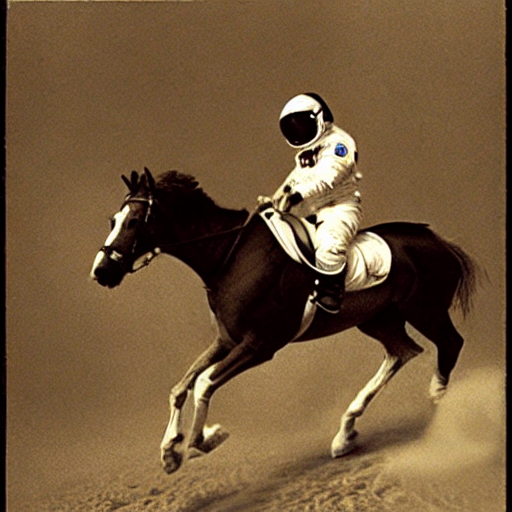}} 
    {\includegraphics[width=0.15\textwidth]{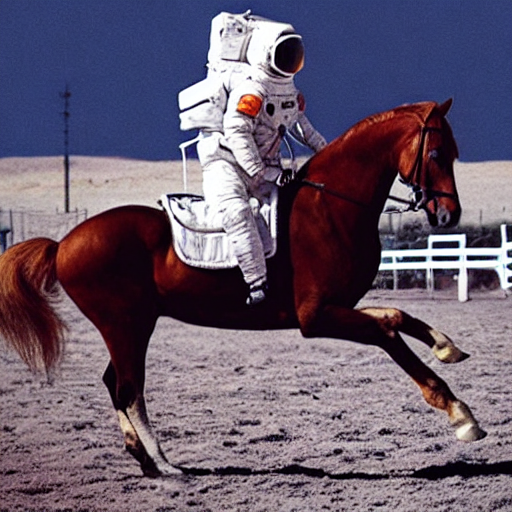}} 
    {\includegraphics[width=0.15\textwidth]{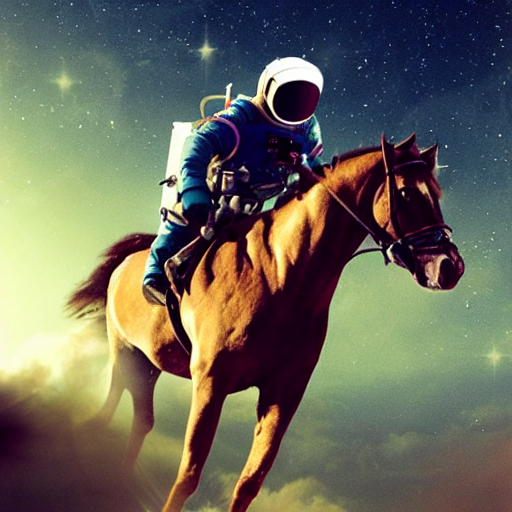}} 
    {\includegraphics[width=0.15\textwidth]{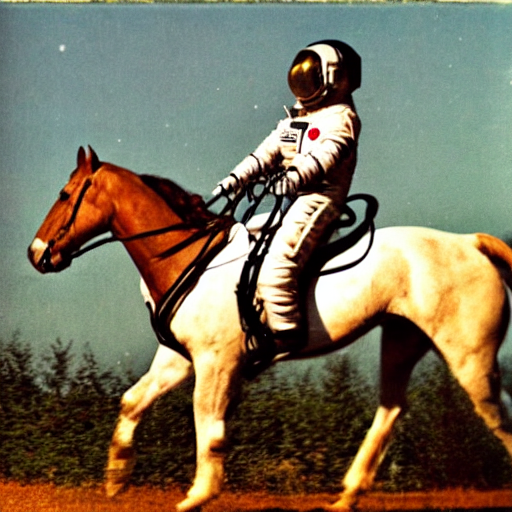}} 
    {\includegraphics[width=0.15\textwidth]{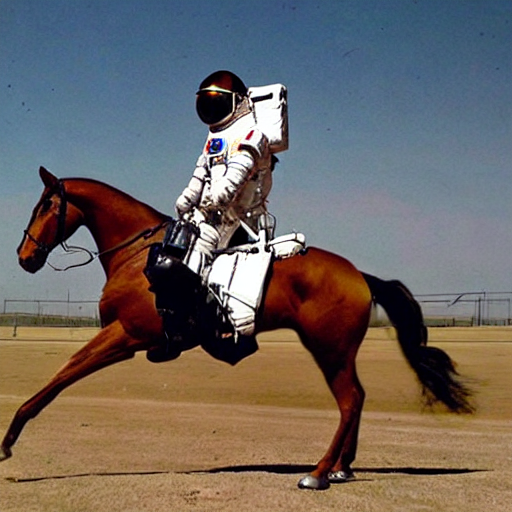}} }
    \\
    \vskip 0.1cm

    \subfloat[Images generated by LDM \emph{after} fine-tuning with our disguises using DreamBooth]{{\includegraphics[width=0.15\textwidth]{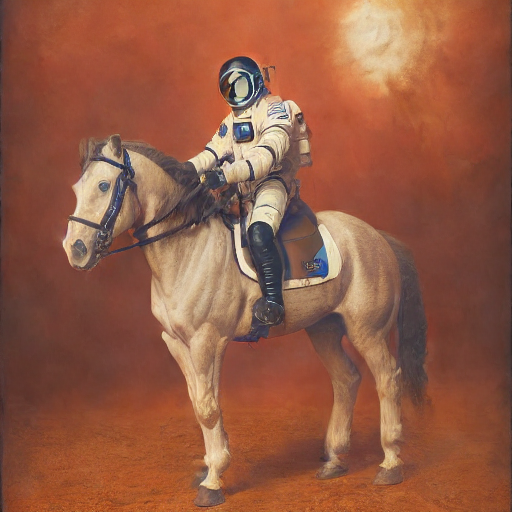}}
    {\includegraphics[width=0.15\textwidth]{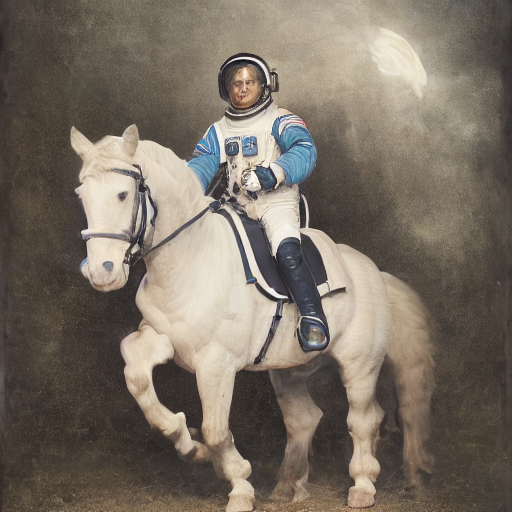}}
    {\includegraphics[width=0.15\textwidth]{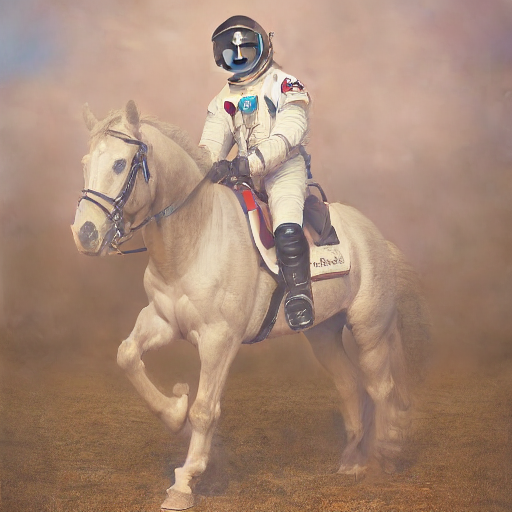}}
    {\includegraphics[width=0.15\textwidth]{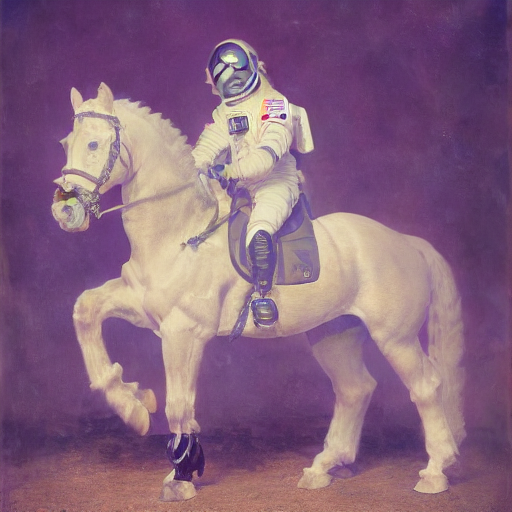}}
    {\includegraphics[width=0.15\textwidth]{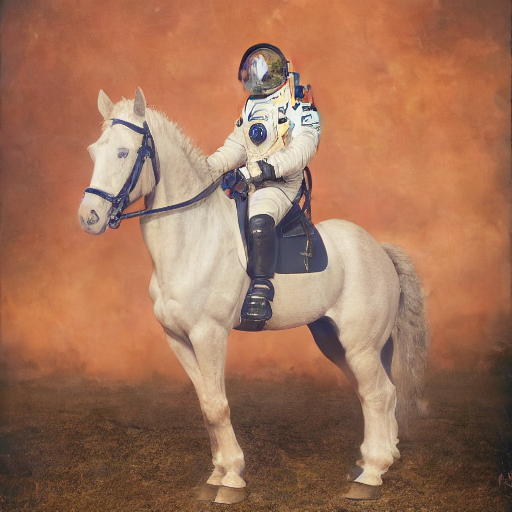}}
    
    {\includegraphics[width=0.15\textwidth]{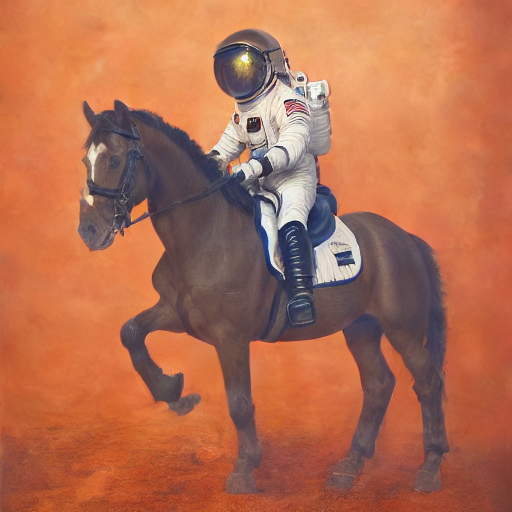}}}\\
     
    \caption{Utility validation: images generated by LDM using the prompt ``a photo of an astronaut riding a horse on Mars'' before/after fine-tuning with our disguises in \Cref{fig:dreambooth} (row 2, column 1) with DreamBooth.}
    \label{fig:utility}
\end{figure}

\subsection{Evaluation on DreamBooth}

\paragraph{Main results:} In \Cref{sec:exp}, we reveal the latent (copyrighted) information contained in the acquired samples $x_d$ using textual inversion, where the diffusion model (i.e., the U-Net in LDM) is not retrained. To further demonstrate the effectiveness of the disguises $x_d$ and examine disguised copyright infringement in other scenarios, we further perform evaluation on DreamBooth \citep{ruiz2023dreambooth}, an LDM-based fine-tuning method on a small set of images designed for subject-driven generation. Specifically, we choose five images from the DreamBooth dataset\footnote{\url{https://github.com/google/dreambooth}}  as the (designated) copyrighted image $x_c$, and generate their corresponding disguises $x_d$ using \Cref{alg:DG} by choosing the noisy version of $x_c$ as the base images $x_b$, training 10000 epochs and setting the weight parameter $\alpha=1000$. After generating the disguises $x_d$, we apply DreamBooth for fine-tuning (on $x_d$ only) and inference  following the general recipe\footnote{\url{https://github.com/huggingface/diffusers/tree/main/examples/dreambooth}}. In \Cref{fig:dreambooth}, we show the copyrighted images $x_c$ in the first row, their corresponding disguises $x_d$ in the second row, and images generated by DreamBooth by fine-tuning with the above $x_d$ in the third row. Our results qualitatively confirm that copyrighted information in $x_c$ can be reproduced by fine-tuning on $x_d$ with DreamBooth.

\paragraph{Evaluation on model utility:} To further examine the practicality of the disguises, we want to demonstrate that the model utility is not deprecated after training (or fine-tuning) on $x_d$. To accomplish this task, we perform the same text-to-image generation task with the prompt
``a photo of an astronaut riding a horse on Mars'' before/after training on $x_d$ with LDM \footnote{We perform inference following the recipe in  \url{https://huggingface.co/CompVis/stable-diffusion-v1-4}}. Specifically, the ``before'' model is Stable Diffusion v1-4, and the ``after'' model is the acquired model by fine-tuning the ``before'' model on the dog disguise (second row, first column in \Cref{fig:dreambooth}). We show our results in \Cref{fig:utility} for 6 inferences and observe that the generation ability of the model is not compromised. We will show in the next section that this observation extends to LDM training.

\subsection{Mixed-training on CelebA-HQ}
\label{sec:mix}
\begin{figure}[htb!]
    \centering
    \subfloat[Target images $x_c$ with the copyrighted symbol]
    {{\includegraphics[width=0.15\textwidth]{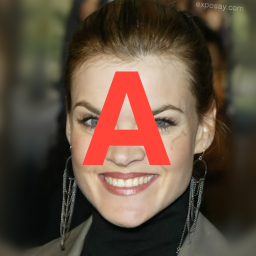}} 
    {\includegraphics[width=0.15\textwidth]{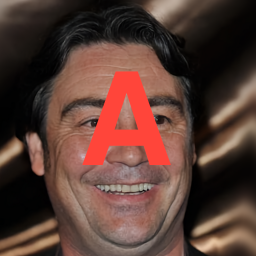}} 
    {\includegraphics[width=0.15\textwidth]{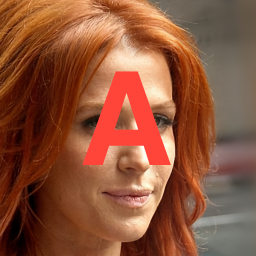}} 
    {\includegraphics[width=0.15\textwidth]{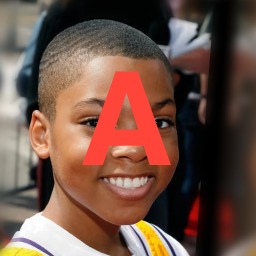}} 
    {\includegraphics[width=0.15\textwidth]{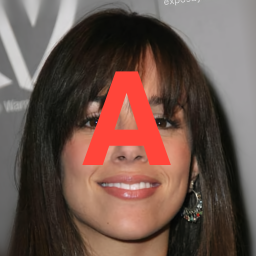}} }
    \\
    \vskip 0.1cm

    \subfloat[Disguises $x_d$]{{\includegraphics[width=0.15\textwidth]{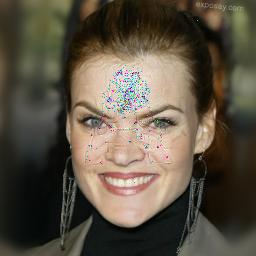}}
    {\includegraphics[width=0.15\textwidth]{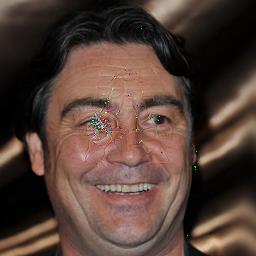}}
    {\includegraphics[width=0.15\textwidth]{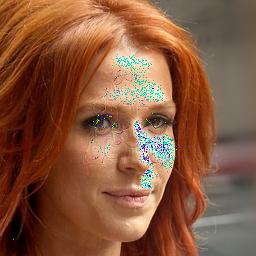}}
    {\includegraphics[width=0.15\textwidth]{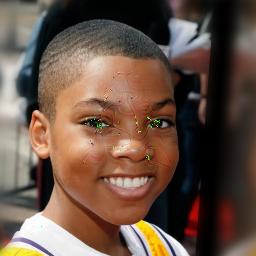}}
    {\includegraphics[width=0.15\textwidth]{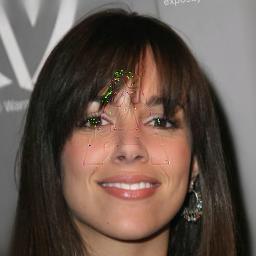}}
    }\\
    \vskip -0.03cm
    
    \subfloat[]{{\includegraphics[width=0.15\textwidth]{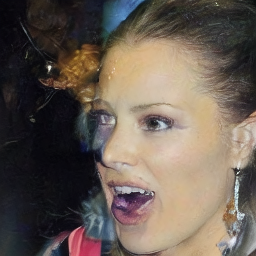}}

    {\includegraphics[width=0.15\textwidth]{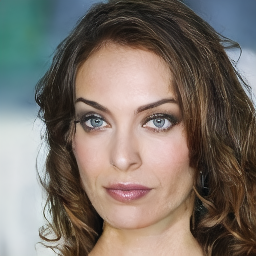}}
    {\includegraphics[width=0.15\textwidth]{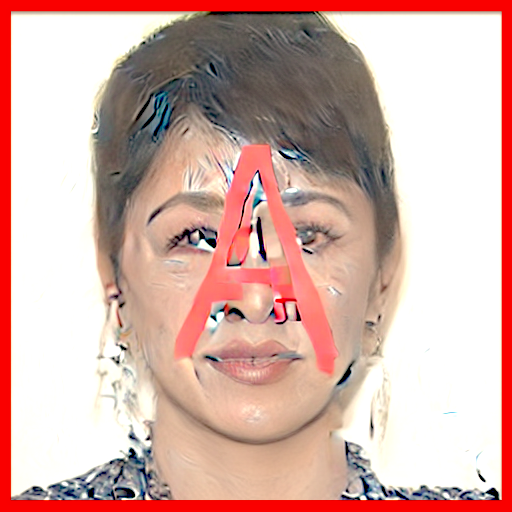}}
    {\includegraphics[width=0.15\textwidth]{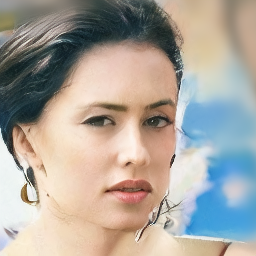}}
    {\includegraphics[width=0.15\textwidth]{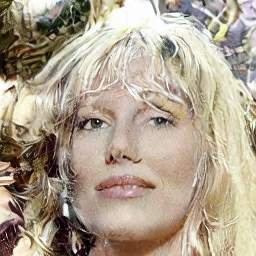}}}\\
    \vskip -0.5cm
    \subfloat[]{
    {\includegraphics[width=0.15\textwidth]{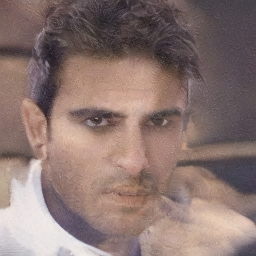}}
    {\includegraphics[width=0.15\textwidth]{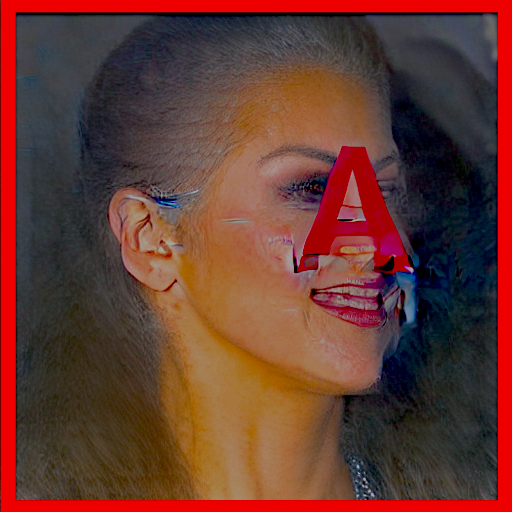}}
    {\includegraphics[width=0.15\textwidth]{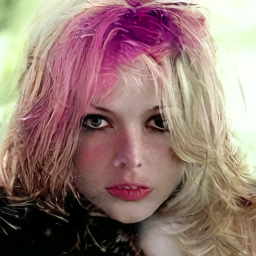}}
    {\includegraphics[width=0.15\textwidth]{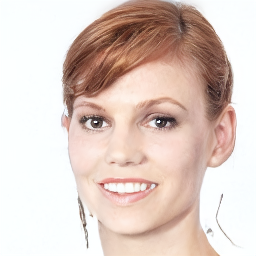}}
    {\includegraphics[width=0.15\textwidth]{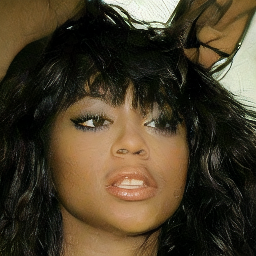}}
    }\\
    \vskip -0.5cm

    \subfloat[Images generated by LDM by training on clean images (1000) and disguises (100)]{
    {\includegraphics[width=0.15\textwidth]{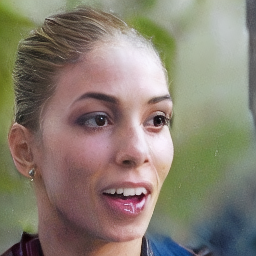}}
    {\includegraphics[width=0.15\textwidth]{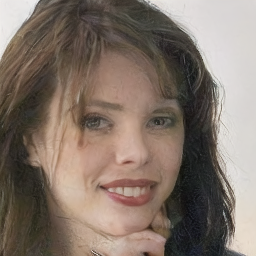}}
    {\includegraphics[width=0.15\textwidth]{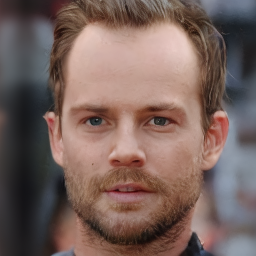}}
    {\includegraphics[width=0.15\textwidth]{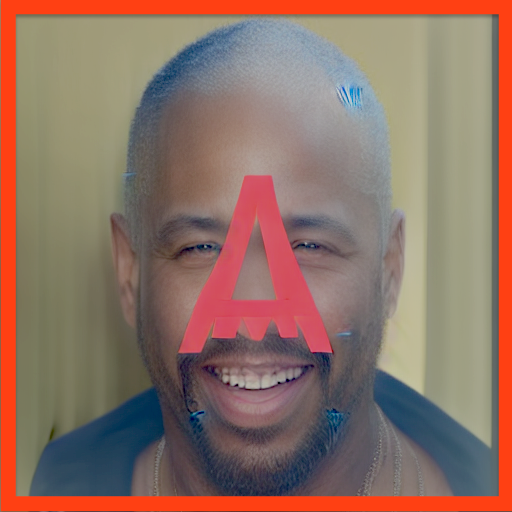}}
    {\includegraphics[width=0.15\textwidth]{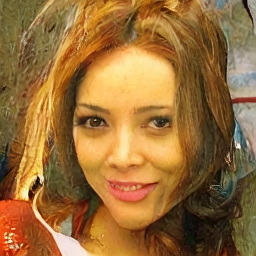}}
    }\\

    \caption{Disguised copyright infringement during mixed-training for unconditional generation on CelebA-HQ. The first row: the designated copyrighted image $x_c$ with a red symbol ``A''; the second row: the corresponding disguises $x_d$; the third to fifth rows: images generated by LDM by training on 1000 clean images and 100 disguises (each disguise above contributes 20 copies). Images highlighted with a red box indicate reproductions of the copyrighted symbol.}
    \label{fig:celeba}
\end{figure}

In the previous experiments, we have performed textual inversion and fine-tuning with DreamBooth where the disguises $x_d$ are the entire training set. In this section, we extend our experiments to more challenging settings where the disguises $x_d$ only constitute a small portion of the training set and are overwhelmed by the clean samples. Specifically, we randomly take a subset of 1000 images from CelebA-HQ/256\footnote{\url{https://www.tensorflow.org/datasets/catalog/celeb_a_hq}} as clean training samples, and another subset of 5 samples (outside of the clean set) and add a (designated) copyrighted symbol ``A'' to all of them to construct $x_c$ (first row in \Cref{fig:celeba}). Next, we construct the corresponding disguises $x_d$ (second row in \Cref{fig:celeba}) to visually remove the symbol using \Cref{alg:DG} and duplicate each disguise 20 times to construct a disguise set containing 100 images. Overall, we acquire a training set of 1100 images where the disguise fraction is $\frac{1}{11}$. We initialize the model with the CompVis official model\footnote{\url{https://huggingface.co/CompVis/ldm-celebahq-256}} to preserve reasonable generation performance and train the model on the mixed dataset for 200 DDIM steps. In \Cref{fig:celeba} (rows 3-5) we show the images generated by the model after training (inference for 20 times) and observe 3 out of 20 images contain the copyrighted symbol, which demonstrated the effectiveness of the disguises for mixed-training scenarios.

\end{document}